\pdfoutput=1

\documentclass[11pt]{article}

\usepackage{acl}

\usepackage{times}
\usepackage{latexsym}

\usepackage[T1]{fontenc}

\usepackage[utf8]{inputenc}

\usepackage{microtype}

\usepackage{inconsolata}

\usepackage{graphicx}

\usepackage{microtype}
\usepackage{hyperref}
\usepackage{url}
\usepackage{booktabs}
\usepackage[nolist]{acronym}                
\usepackage{adjustbox}
\usepackage{amsmath}
\usepackage{amssymb}
\usepackage{graphicx}
\usepackage{cleveref}
\usepackage{dirtytalk}
\usepackage{tcolorbox}
\usepackage{subcaption}
\usepackage{wrapfig}
\usepackage{multicol}
\usepackage{multirow}
\usepackage{makecell}
\usepackage{longtable}

\definecolor{darkblue}{rgb}{0, 0, 0.5}
\hypersetup{colorlinks=true, citecolor=darkblue, linkcolor=darkblue, urlcolor=darkblue}

\newcommand{\lang}[1]{\texttt{\MakeLowercase{#1}}}

\title{LawInstruct: A Resource for Studying Language Model Adaptation\\to the Legal Domain}

\author{
Joel Niklaus $^{S\,X}$
\,
Lucia Zheng $^{S}$
\,
Arya D. McCarthy $^{J}$
\,
Christopher Hahn $^{X}$
\,
Brian M. Rosen $^{X}$
\AND
Peter Henderson $^{S\,P\,\thanks{equal co-supervision in alphabetical order}}$
\,
Daniel E. Ho $^{S\,^*}$
\,
Garrett Honke $^{X\,^*}$
\,
Percy Liang $^{S\,^*}$
\,
Christopher Manning $^{S\,^*}$
\\\\
$^S$Stanford University\,
$^J$Johns Hopkins University\,
$^X$X, the Moonshot Factory\,
$^P$Princeton University\\
}

\begin{document}

\maketitle
\begin{abstract}
Instruction tuning is an important step in making language models useful for direct user interaction. However, the legal domain is underrepresented in typical instruction datasets (e.g., only 10 out of 1600+ tasks in Super-NaturalInstructions). To study whether instruction tuning on legal datasets is necessary for strong legal reasoning, we aggregate 58 annotated legal datasets and write instructions for each, creating LawInstruct. LawInstruct covers 17 global jurisdictions, 24 languages and a total of 12M examples across diverse tasks such as legal QA, summarization of court cases, and legal argument mining. We evaluate our models on LegalBench, measuring legal reasoning across five categories in 162 challenging and realistic legal tasks, and MMLU, to measure potential drops in general reasoning capabilities. We find that legal-specific instruction tuning on Flan-T5 – yielding FLawN-T5 – improves performance on LegalBench across all model sizes, with an aggregate increase of 15 points or 50\% over Flan-T5 for the base size. No model size shows performance drops in MMLU. We publish LawInstruct as a resource for further study of instruction tuning in the legal domain.
\end{abstract}

\section{Introduction}
\label{sec:introduction}
In recent years, Large Language Models (LLMs) advanced significantly, evident in their performance gains across numerous benchmarks, including SuperGLUE \citep{wang_superglue_2019}, MMLU \citep{hendrycks_measuring_2021}, and various human examinations~\citep{openai_gpt-4_2023}, such as the U.S. bar exams for law practice admission~\citep{katz_gpt-4_2023}.
However, the interplay between domain-specific training and within-domain evaluation is poorly understood. This work examines how training on domain-specific legal corpora affects performance on the widest set of legal-domain evaluation benchmarks known to the authors. 
We thus conduct a study of the ability of models to answer questions, classify, make judgments, extract information, and otherwise perform decision making or higher-order cognitive tasks (i.e., to ``reason'') within a limited domain, as opposed to broad-domain benchmarking.
We present evidence that domain-specific pretraining and instruction tuning improve performance---but the effect does not generalize across all tasks, training regimes, model sizes, and other factors. 

Although large closed models also still hallucinate heavily on legal texts \citep{dahl_large_2024}, they achieve much better performance on LegalBench than smaller open models (e.g., 77.3 for GPT-4 vs.\ 60.1 for Flan-T5 XXL, the state-of-the-art open model). 
In the legal domain it is often crucial for reasons of trust and data protection not to use public models, so many firms need on-premise deployments. Therefore models like Claude or GPT-4 cannot be used, stressing the need for open models. In this study, we explore the potential of enhancing model performance through in-domain instruction tuning and continued pretraining on Flan-T5, the current state-of-the-art open model on LegalBench in both the 3B and 11B range.

\begin{figure*}[t]
\vspace{-3mm}
    \centering
    \includegraphics[width=\textwidth]{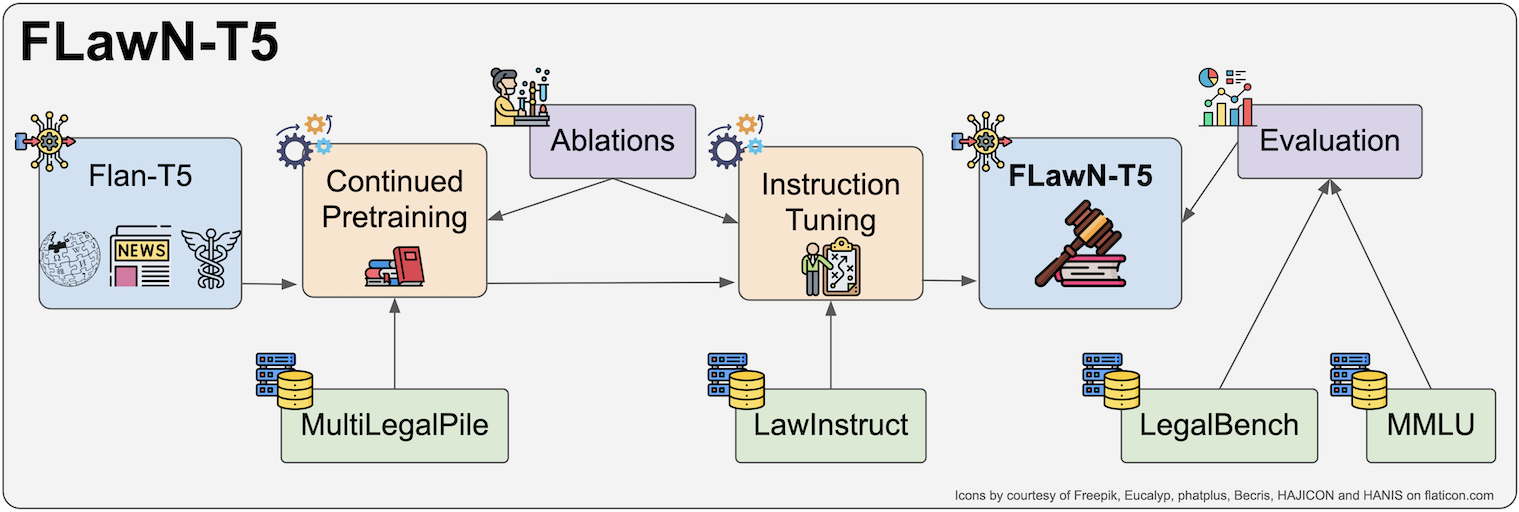}
    \vspace{-6mm}
    \caption{We continue pretraining on MultiLegalPile, instruction tune on LawInstruct and evaluate on LegalBench and MMLU.}
    \label{fig:flawn_t5}
\vspace{-3mm}
\end{figure*}

To study this, we use the MultiLegalPile~\citep{niklaus_multilegalpile_2024}, a 689GB multilingual legal corpus, for continued pretraining. Because no instruction dataset for legal reasoning is available, we introduce LawInstruct, spanning 24 languages in 17 jurisdictions on four continents. It contains 12M training examples for QA, entailment, summarization, and information extraction tasks in the legal domain, each presented as a bespoke instruction with corresponding output. With this large instruction dataset in hand, we fine-tune models and then perform quantitative analyses of their outputs on the LegalBench \citep{guha_legalbench_2023} and MMLU \citep{hendrycks-etal-2021-measuring} benchmark suites. 
Instruction tuning Flan-T5 models on LawInstruct, we achieve a balanced accuracy of 58.1 on LegalBench for the XL size, improving by 8 points or 16\% over the baseline. The Small model even improves by 9.6 points or 38.1\% and by 14 points or 55.4\% when we also continue pretraining it. 



The contributions of this paper are four-fold:
First, we curate the first legal instruction dataset by standardizing and writing instructions for 58 high-quality annotated datasets covering diverse legal tasks to make them usable for instruction tuning in the first place.
Second, we continue pretraining and instruction tune T5, mT5, and Flan-T5 models and achieve new state-of-the-art on LegalBench in all tested parameter ranges.
Third, we perform a wide range of ablations across different dataset configurations deepening our understanding of adapting models to specific domains.
Finally, we publicly release the permissively-licensed portion of the curated dataset on the Hugging Face Hub\footnote{\url{https://huggingface.co/lawinstruct/lawinstruct}} and release the code used to create the dataset\footnote{\url{https://github.com/JoelNiklaus/LawInstruct/}} including pointers on how to access the portions of the data that require special agreements.
\section{Experimental Setup}
In this section, we describe the experimental setup we used to test the effect of pretraining and instruction tuning on in-domain legal data. 
We use random seed 42 throughout. Our experiments were performed with T5X \footnote{\url{https://github.com/google-research/t5x}} on TPUv4 pods using 2 to 512 cores. We present the mean across tasks per LegalBench category and for LegalBench overall by aggregating over the categories.
We consider T5 v1.1+LM adaptation \citep{raffel_exploring_2020, lester_power_2021}, Flan-T5 \citep{chung_scaling_2022} and mT5 \citep{xue_mt5_2021} models in the sizes Small, Base, XL and XXL, allowing us to study effects over different model scales. We selected the T5 family of models over other models for three reasons: 1) Flan-T5 XL and XXL perform best in their parameter range on LegalBench, 2) T5 and mT5 allow us to measure the effect of multilinguality in a controlled setting, and 3) the T5 model family contains models from 60M parameters (Small) to 11B (XXL) allowing us to study scaling behaviour also at smaller scales.



\subsection{Continued Pretraining}

We continue pretraining on the \mbox{\textbf{MultiLegalPile}}~\citep{niklaus_multilegalpile_2024}, a 689GB corpus in 24 languages from 17 jurisdictions. It includes diverse legal data sources with varying licenses and allows for pretraining NLP models under fair use, with more permissive licenses for the Eurlex Resources and Legal mC4 subsets. It consists of four large subsets: a) Native Multi Legal Pile (112 GB), b) Eurlex Resources (179 GB), c) Legal mC4 (106 GB), and d) Pile of Law (292 GB). For our mT5 experiments, we use the entire corpus, and for T5 and Flan-T5 experiments, we use only English texts.

We continued pretraining (a.k.a.\ domain adaptation of) with 512 tokens in both inputs and targets on the MultiLegalPile \citep{niklaus_multilegalpile_2024} whereas the original models were pretrained on C4 \citep{raffel_exploring_2020}. We used the UL2 mixture \citep{tay_unifying_2022} due to its promise to enable improved training efficiency with its mixture of denoisers.
In initial experiments we used batch size 1024 and warmed up the learning rate linearly for the first 10K steps from 2.5e-3 to 5e-3, then decayed it to 1.5e-3. However, we noticed training instabilities for the XXL models. We switched to a constant learning rate of 1e-3 and ran a sweep over batch sizes 64, 128, 256, 512, 1024. The XXL model trained stably only with batch size 128.

\subsection{Instruction Tuning}

\begin{figure*}[ht]
\vspace{-3mm}
    \centering
    \includegraphics[width=\textwidth]{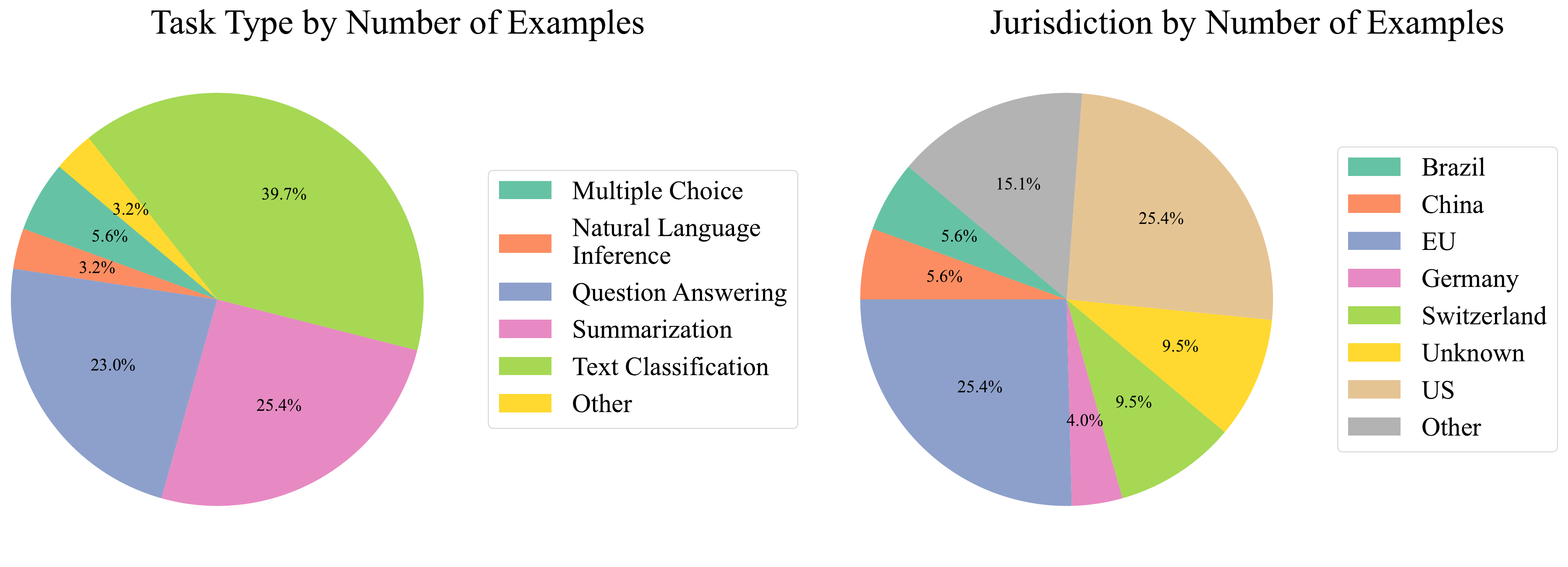}
    \vspace{-12mm}
    \caption{Jurisdiction and task type by examples.}
    \label{fig:lawinstruct_by_examples}
\vspace{-3mm}
\end{figure*}

\begin{figure*}[ht]
\centering
\begin{subfigure}[b]{0.33\textwidth}
    \centering
    \includegraphics[width=\columnwidth]{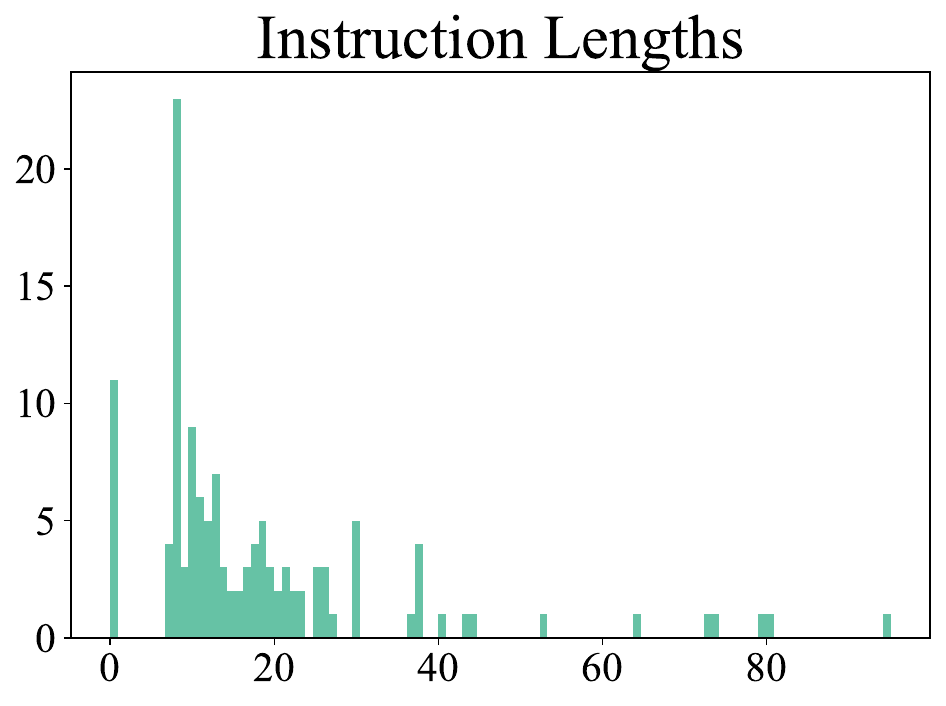}
    \caption{Instructions (capped at 100)}
    \label{fig:instruction_length_histogram}
\end{subfigure}%
\hfill 
\begin{subfigure}[b]{0.33\textwidth}
    \centering
    \includegraphics[width=\columnwidth]{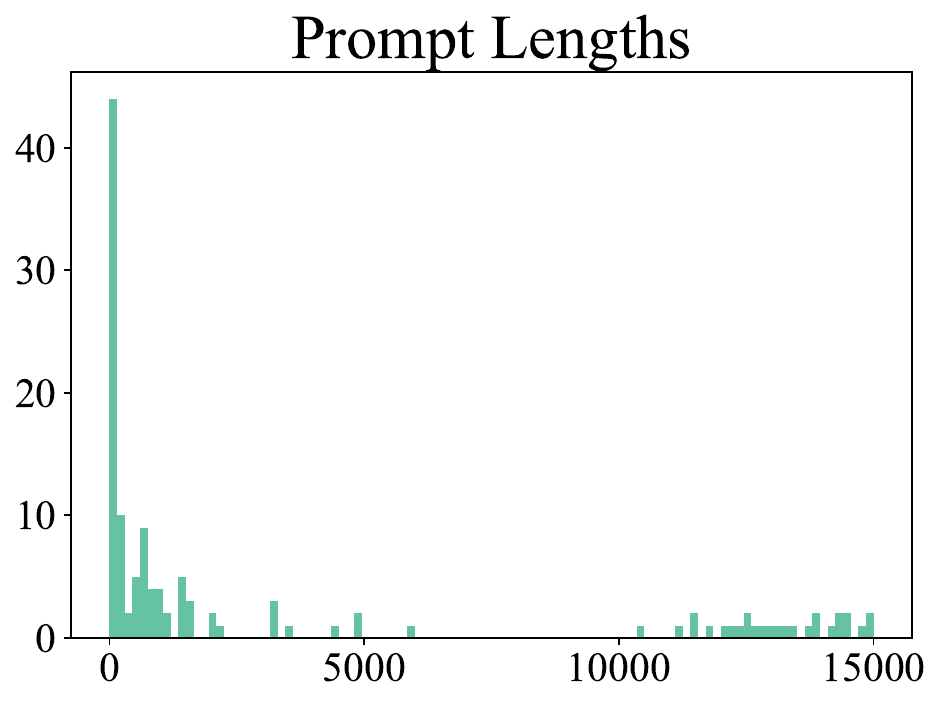}
    \caption{Prompts (capped at 15K)}
    \label{fig:prompt_length_histogram_capped_at_15000}
\end{subfigure}
\hfill 
\begin{subfigure}[b]{0.33\textwidth}
    \centering
    \includegraphics[width=\columnwidth]{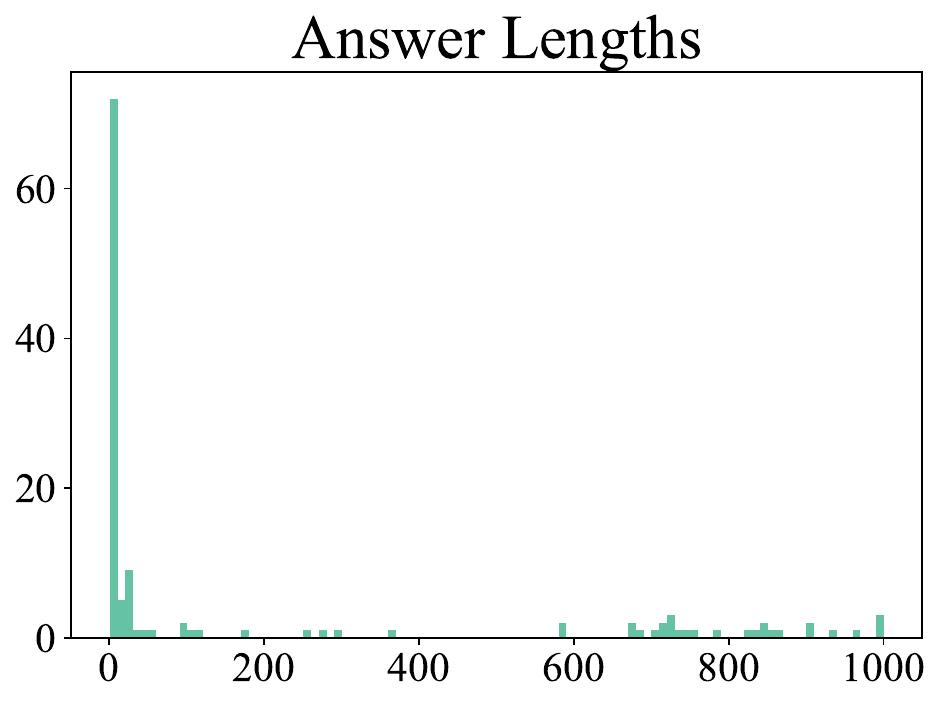}
    \caption{Answers (capped at 1K)}
    \label{fig:answer_length_histogram_capped_at_1000}
\end{subfigure}
\vspace{-6mm}
\caption{Mean length distributions for instructions, prompts and answers.}
\label{fig:lawinstruct_length_distribution}
\vspace{-3mm}
\end{figure*}

In this paper, we are interested in the ability of LLMs to answer questions, make judgments, and perform decision making (i.e., to ``reason'') within the legal domain. Legal reasoning is often highly sensitive, and the struggles of factuality in LLMs lead to legalese with \say{bogus judicial decisions, bogus quotes, and bogus internal citations} \citep{nyt,dahl_large_2024}. 
In the absence of legal instruction datasets and to evaluate the effect of legal instruction tuning on models' capability to reason in legal domains, we develop \textbf{LawInstruct}: a large instruction dataset that normalizes and adapts 58 existing or novel legal-domain datasets with custom templates. LawInstruct is the first instruction dataset in the legal domain known to the authors. We attempted to collect a broad sample of datasets to expose the model to a variety of legal systems and concepts. 
We started by taking the datasets operating on legal data from Natural Instructions \citep{mishra_cross-task_2022, wang_super-naturalinstructions_2022} and then surveyed the literature to select high-quality legal datasets. The resulting dataset contains a total of almost 12M 
examples in 24 languages.
Data sources and detailed statistics including license, language and jurisdiction are given in \Cref{sec:detailed_dataset_description} \Cref{tab:dataset}. 
Each example is built from a human-written task-specific template: We write a simple instruction per task (107 in total), we take the input of the supervised dataset as the prompt and the output as the answer (see \Cref{fig:example} for an example). 
We show pie charts visualizing the composition of LawInstruct across the jurisdiction and task types calculated by the number of examples in \Cref{fig:lawinstruct_by_examples} and by datasets in \Cref{sec:detailed_dataset_description} \Cref{fig:lawinstruct_by_datasets}. \Cref{fig:lawinstruct_length_distribution} shows the length distribution of the instructions, prompts and answers following \citet{wang-etal-2023-camel}.


\begin{figure}
\vspace{-3mm}
    \centering
    \begin{tcolorbox}[left=1mm,right=1mm,top=1mm,bottom=1mm,middle=1mm,colback=white!90!black]
    \small \color{black}
\textit{Instruction:} \textcolor{blue}{Answer the following legal question. Cite relevant evidence when possible.}
\\\\
\textit{Prompt:} \textcolor{blue}{Question: So say a bank robber walks into a bank and hands the teller a note saying simply "please give me \$1,000" (or maybe even just verbally saying "I need \$1,000 please") and if the teller accepts, which they probably would because they're supposed to comply with the robber, he takes the money, otherwise if they refuse, he leaves. How is this different than say people asking for charity donations in the street? He just politely requested money without making any threats.}
\\\\
\textit{Answer:} \textcolor{blue}{Answer: As a concrete example, consider Missouri v. Coleman, where Coleman handed a teller a plastic bag and said "I need you to do me a favor. Put the money in this bag", and later as the manager approached, said "Ma'am, stop where you are and don't move any farther". Coleman was convicted of second degree robbery: but the appeals court found that he had not acted forcibly, so his conviction was overturned. Instead, the court entered a conviction for the lesser offense of stealing which is when one appropriates property or services of another with the purpose to deprive him or her thereof, either without his or her consent or by means of deceit or coercion. The act may also be termed "theft", as in Washington state. The thief is acting deceptively and thereby gaining control over property.}
    \end{tcolorbox}
    \vspace{-3mm}
    \caption{Instruction template in LawInstruct for StackExchangeQuestionsLegal populated with \emph{instruction}, \emph{prompt} and \emph{answer}. Models are trained to generate the answer conditioned on the instruction and prompt.}
    \label{fig:example}
\vspace{-3mm}
\end{figure}

We finetuned the models with 2048 input and 512 target tokens. We ran a hyperparameter sweep for the XL model over the learning rate (5e-5, 1e-5, 5e-4, 1e-4, 5e-4) and dropout (0, 0.05, 0.1, 0.15, 0.2, 0.25), with learning rate 5e-4 and dropout 0.15 achieving the best validation loss. Unless specified otherwise, we trained the models for 2K steps with batch size 64.
In addition to LawInstruct, we used an updated Flan mixture \citep{chung_scaling_2022}. 
We built the input by concatenating the prompt with two new lines, the instruction and two additional new lines. Per LawInstruct config, we used the first 16 examples for validation and the remaining ones for training. We selected the model with the best LawInstruct validation loss. 
While in-context learning has achieved strong results in many tasks \citep{brown_language_2020}, further finetuning language models for specific tasks may still be necessary for better results \citep{mosbach_few-shot_2023}.

\subsection{Evaluation}

We evaluate our models on LegalBench and MMLU to test in-domain and generalization performance, respectively.
\textbf{LegalBench}~\citep{guha_legalbench_2023} consists of 162 tasks evaluating different aspects of legal classification and reasoning. Each task is assigned to one of five categories, depending on the broader type of legal reasoning implicated. LegalBench tasks are sourced from both previously constructed datasets and novel tasks collected from different members of the legal community (e.g., lawyers, legal impact organizations, legal academics). As such, LegalBench is thought to capture tasks of interest and practical applicability. LegalBench tasks span a wide range of legal subject areas (e.g., contracts, civil procedure, tax, etc.) and text-types (natural language, contractual terms, judicial opinions, etc.). The majority of tasks are either classification or extraction tasks, thus enabling automated evaluation.
Massively Multilingual Language Understanding (\textbf{MMLU}) benchmarks models factual knowledge \citep{hendrycks-etal-2021-measuring}. MMLU contains multiple-choice questions on 57 subjects, including three related to law: jurisprudence, international law, and professional law.
We conducted preliminary experiments on the multilingual LEXTREME benchmark \citep{niklaus_lextreme_2023} but encountered significant obstacles: a) Many tasks in LEXTREME are NER-based, posing evaluation challenges for generative models like ours and b) The extreme multi-label classification tasks in LEXTREME showed near-random performance without task-specific fine-tuning, which is beyond the scope of our current work. Therefore, we focus on LegalBench and MMLU, both in English.

For evaluation, we set temperature to 0 in line with accepted practice for LegalBench evaluation \citep{guha_legalbench_2023} that focuses on the highest-likelihood token sequence with minimal variance. We removed the following prefixes before scoring: ``label'', ``target'', ``option'', ``answer'', ``a:''. We did not evaluate Rule QA because it necessitated manual evaluation.
We show paper baseline results compared with our runs in \Cref{sec:detailed_results} \Cref{tab:paper_baselines}. Our XL model is quite close to the XL model in the LegalBench paper, but there are significant differences for the XXL model. We provide a more detailed analysis of possible causes in \Cref{sec:inexplicable_xxl}. Unless specifically mentioned, we compare to our baselines results. We hold out LegalBench tasks overlapping with LawInstruct tasks unless specified otherwise (see \Cref{sec:detailed_evaluation} for details).





\section{Results}



This section discusses the main results from instruction tuning and continued pretraining Flan-T5.

\begin{figure*}[ht]
\vspace{-3mm}
\centering
\begin{subfigure}[b]{0.5\textwidth}
    \centering
    \includegraphics[width=\columnwidth]{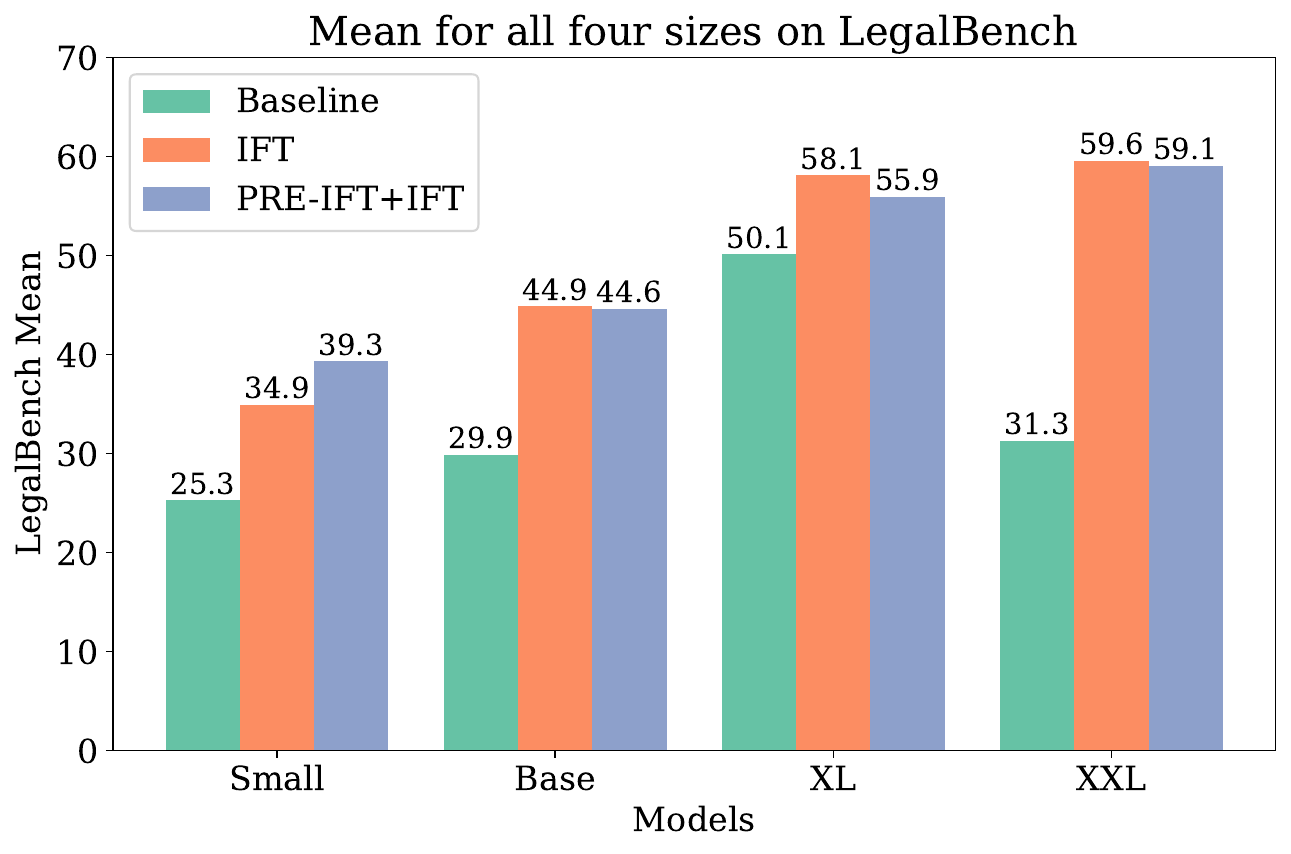}
    \caption{LegalBench}
    \label{fig:progression_lb}
\end{subfigure}%
\hfill 
\begin{subfigure}[b]{0.5\textwidth}
    \centering
    \includegraphics[width=\columnwidth]{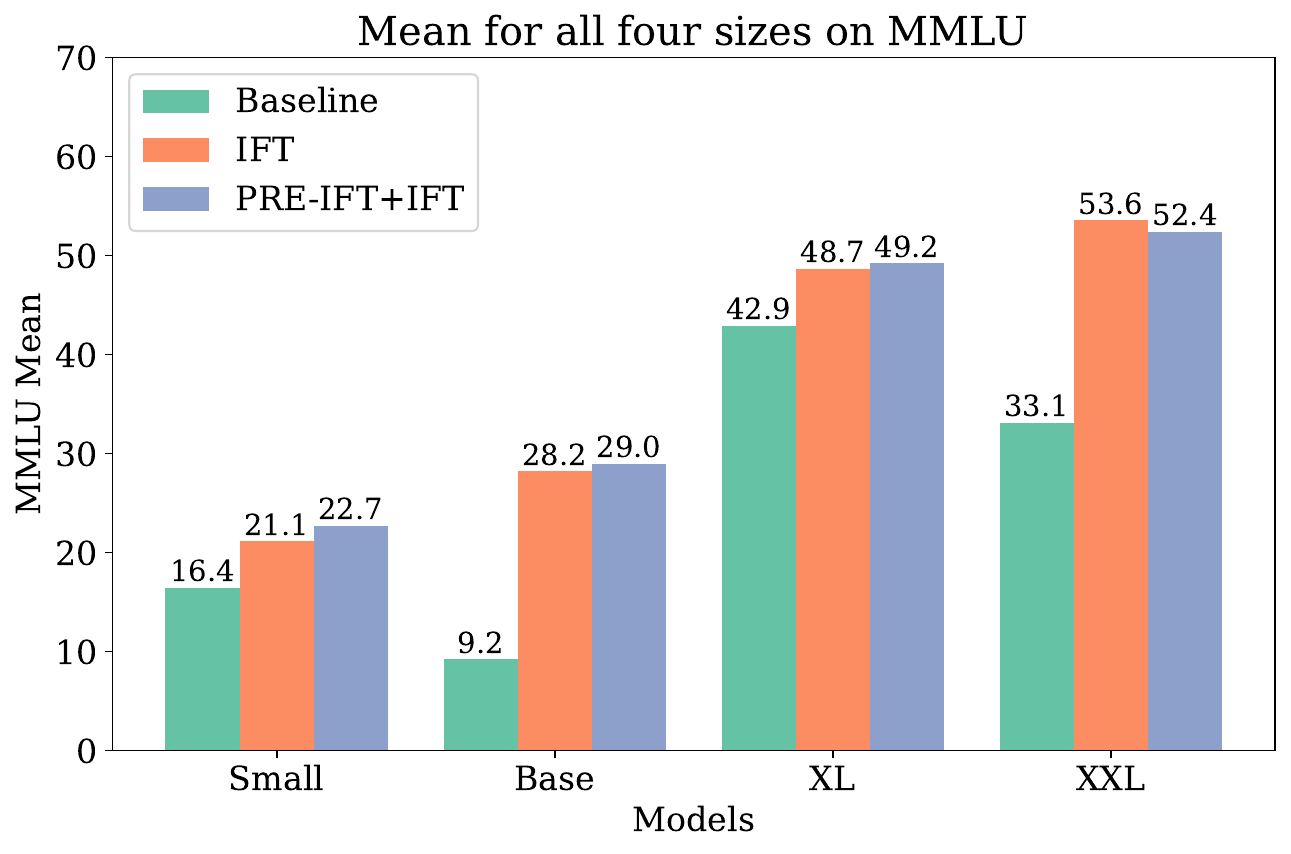}
    \caption{MMLU}
    \label{fig:progression_mmlu}
\end{subfigure}
\vspace{-6mm}
\caption{Performance progression on LegalBench and MMLU from baseline to instruction tuning (IFT) and continued pretraining followed by instruction tuning (PRE-IFT+IFT).}
\label{fig:progression}
\vspace{-3mm}
\end{figure*}

\begin{table*}[ht]
\centering
\caption{Progression of performance from baseline to instruction tuning (IFT) and continued pretraining followed by instruction tuning (PRE-IFT+IFT).}
\resizebox{\textwidth}{!}{
\begin{tabular}{lrrrrrrr}
\toprule
\bf LLM & \bf Issue & \bf Rule & \bf Conclusion & \bf Interpretation & \bf Rhetorical & \bf LegalBench & \bf Improvement \\
\midrule
Small Baseline & 0.3 {\small ± 0.7} & 30.4 {\small ± 20.3} & 39.8 {\small ± 20.8} & 28.2 {\small ± 21.6} & 27.7 {\small ± 21.9} & 25.3 {\small ± 14.8} & - \\
Small IFT & 25.0 {\small ± 22.0} & 38.1 {\small ± 25.4} & 43.0 {\small ± 17.1} & 36.1 {\small ± 26.5} & 32.6 {\small ± 24.2} & 34.9 {\small ± 6.7} & 9.6 (38.1\%) \\
Small PRE-IFT+IFT & 51.6 {\small ± 2.7} & 37.7 {\small ± 25.2} & 39.8 {\small ± 18.4} & 33.7 {\small ± 23.3} & 33.8 {\small ± 22.4} & 39.3 {\small ± 7.4} & 14.0 (55.4\%) \\
\midrule
Base Baseline & 44.7 {\small ± 12.4} & 18.0 {\small ± 23.6} & 20.9 {\small ± 24.8} & 28.9 {\small ± 21.2} & 37.0 {\small ± 21.3} & 29.9 {\small ± 11.1} & - \\
Base IFT & 50.3 {\small ± 2.4} & 38.8 {\small ± 25.9} & 40.5 {\small ± 15.7} & 49.5 {\small ± 19.1} & 45.2 {\small ± 22.0} & 44.9 {\small ± 5.2} & 15.0 (50.2\%) \\
Base PRE-IFT+IFT & 51.6 {\small ± 4.8} & 38.2 {\small ± 25.5} & 44.0 {\small ± 13.4} & 45.4 {\small ± 16.5} & 44.1 {\small ± 19.0} & 44.6 {\small ± 4.8} & 14.8 (49.5\%) \\
\midrule
XL Baseline & 53.5 {\small ± 6.0} & 32.1 {\small ± 24.6} & 46.8 {\small ± 15.6} & 58.7 {\small ± 21.3} & 59.6 {\small ± 25.6} & 50.1 {\small ± 11.3} & - \\
XL IFT & 65.7 {\small ± 15.2} & 45.1 {\small ± 30.3} & 49.5 {\small ± 14.2} & 61.7 {\small ± 17.1} & 68.6 {\small ± 24.1} & 58.1 {\small ± 10.3} & 8.0 (16.0\%) \\
XL PRE-IFT+IFT & 60.3 {\small ± 10.6} & 44.3 {\small ± 29.7} & 50.5 {\small ± 15.4} & 57.3 {\small ± 15.9} & 67.3 {\small ± 23.1} & 55.9 {\small ± 8.9} & 5.8 (11.6\%) \\
\midrule
XXL Baseline & 36.1 {\small ± 21.5} & 18.8 {\small ± 24.6} & 25.2 {\small ± 26.0} & 35.1 {\small ± 22.2} & 41.1 {\small ± 18.4} & 31.3 {\small ± 9.1} & - \\
XXL IFT & 55.2 {\small ± 23.7} & 46.3 {\small ± 31.6} & 56.2 {\small ± 18.3} & 66.3 {\small ± 19.7} & 73.8 {\small ± 24.4} & 59.6 {\small ± 10.6} & 28.3 (90.5\%) \\
XXL PRE-IFT+IFT & 52.2 {\small ± 14.7} & 47.4 {\small ± 30.8} & 59.2 {\small ± 18.3} & 66.6 {\small ± 18.5} & 70.0 {\small ± 24.1} & 59.1 {\small ± 9.5} & 27.8 (89.0\%) \\
\midrule
GPT-4 \citet{guha_legalbench_2023} & 82.9 & 59.2 & 89.9 & 75.2  & 79.4  & 77.3  & - \\
\bottomrule
\end{tabular}
}
\label{tab:progression}
\vspace{-3mm}
\end{table*}

\begin{figure*}[ht]
\vspace{-6mm}
\centering
\begin{subfigure}[b]{0.5\linewidth}
    \centering
    \includegraphics[height=11cm]{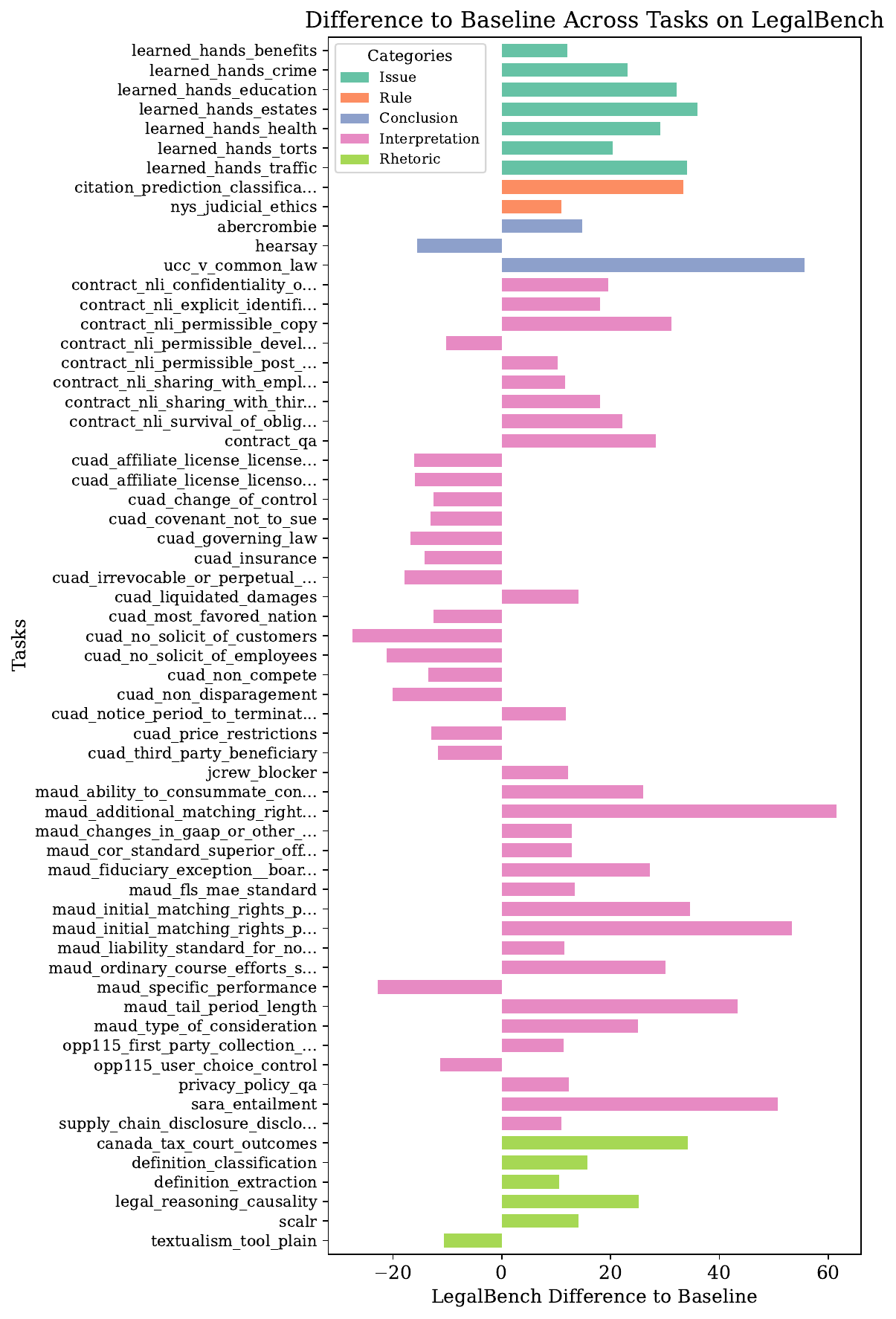}
    \caption{LegalBench}
    \label{fig:difference_to_baseline_tasks_lb}
\end{subfigure}%
\hspace{-8mm}
\begin{subfigure}[b]{0.5\linewidth}
    \centering
    \includegraphics[height=11cm]{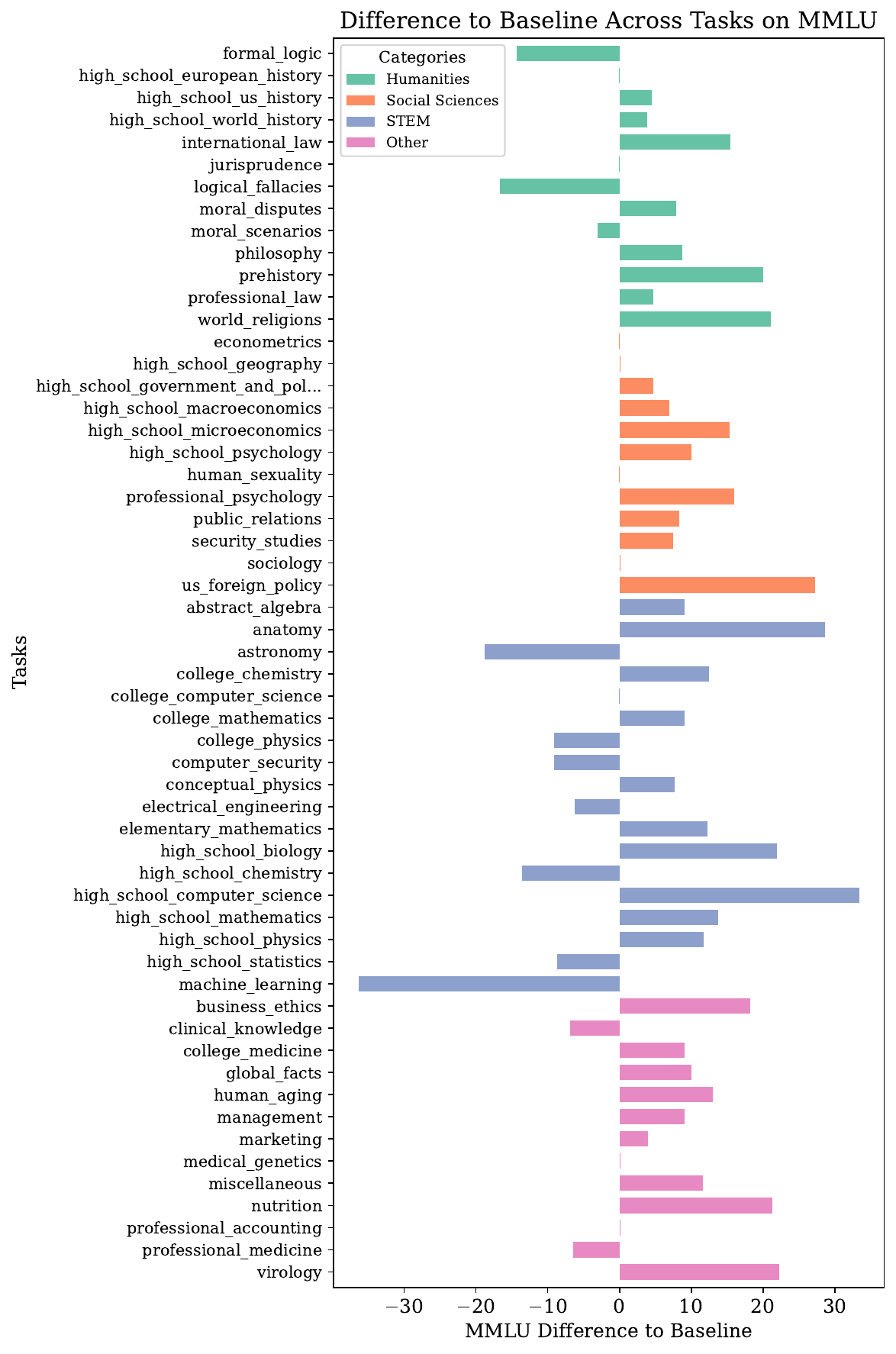}
    \caption{MMLU}
    \label{fig:difference_to_baseline_tasks_mmlu}
\end{subfigure}
\vspace{-3mm}
\caption{Difference to the baseline for the XL model across tasks on LegalBench and MMLU. For LegalBench, we excluded tasks with a difference between -10 and 10 for clarity.}
\label{fig:difference_to_baseline_tasks}
\vspace{-3mm}
\end{figure*}

\Cref{fig:progression} and \Cref{tab:progression} show the performance progression from the baseline over instruction tuning to domain adaptation + instruction tuning on LegalBench and MMLU. Instruction tuning leads to a large performance increase for all model sizes (38.1\% for Small, 50.2\% for Base, 16\% for XL, and 90.5\% for XXL). Domain adaptation + instruction tuning only improves further for the Small model size (55.4\% vs.\ 38.1\%). It seems like larger models benefit less from in-domain pretraining than smaller models, possibly because they can ``remember'' more from the pretraining phase due to increased capacity. Alternatively, a reason for non-consistent improvements of domain adaptation could be the switch from the UL2 tasks in continued pretraining to standard next-token prediction in instruction tuning. Finally, we conjecture that the switch from input length 512 tokens in continued pretraining to 2048 tokens in instruction tuning could have led lower performance for domain-adapted models.

\begin{figure*}[ht]
\vspace{-3mm}
\centering
\begin{subfigure}[b]{0.5\textwidth}
    \centering
    \includegraphics[width=\columnwidth]{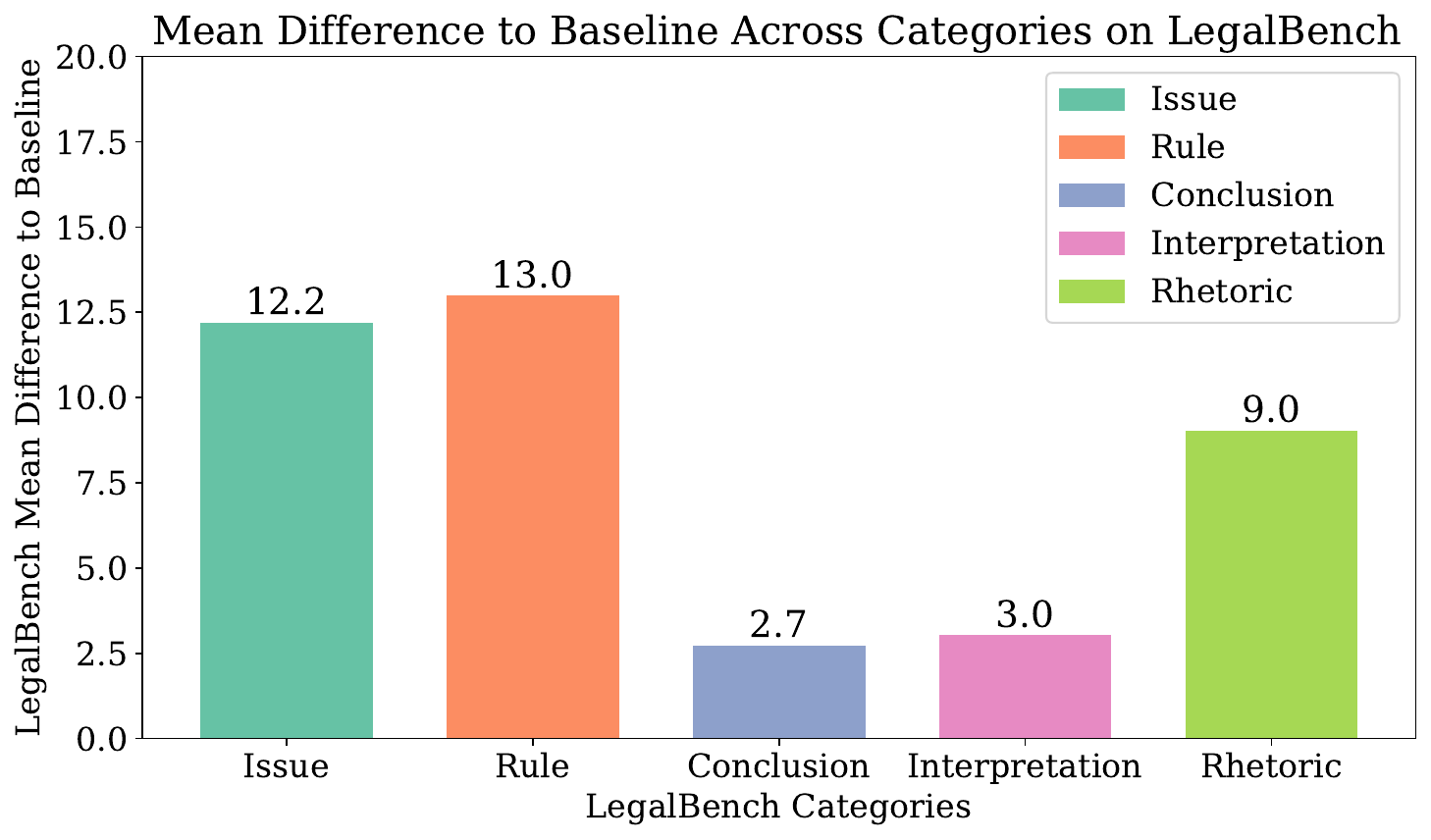}
    \caption{LegalBench}
    \label{fig:difference_to_baseline_categories_lb}
\end{subfigure}%
\hfill 
\begin{subfigure}[b]{0.5\textwidth}
    \centering
    \includegraphics[width=\columnwidth]{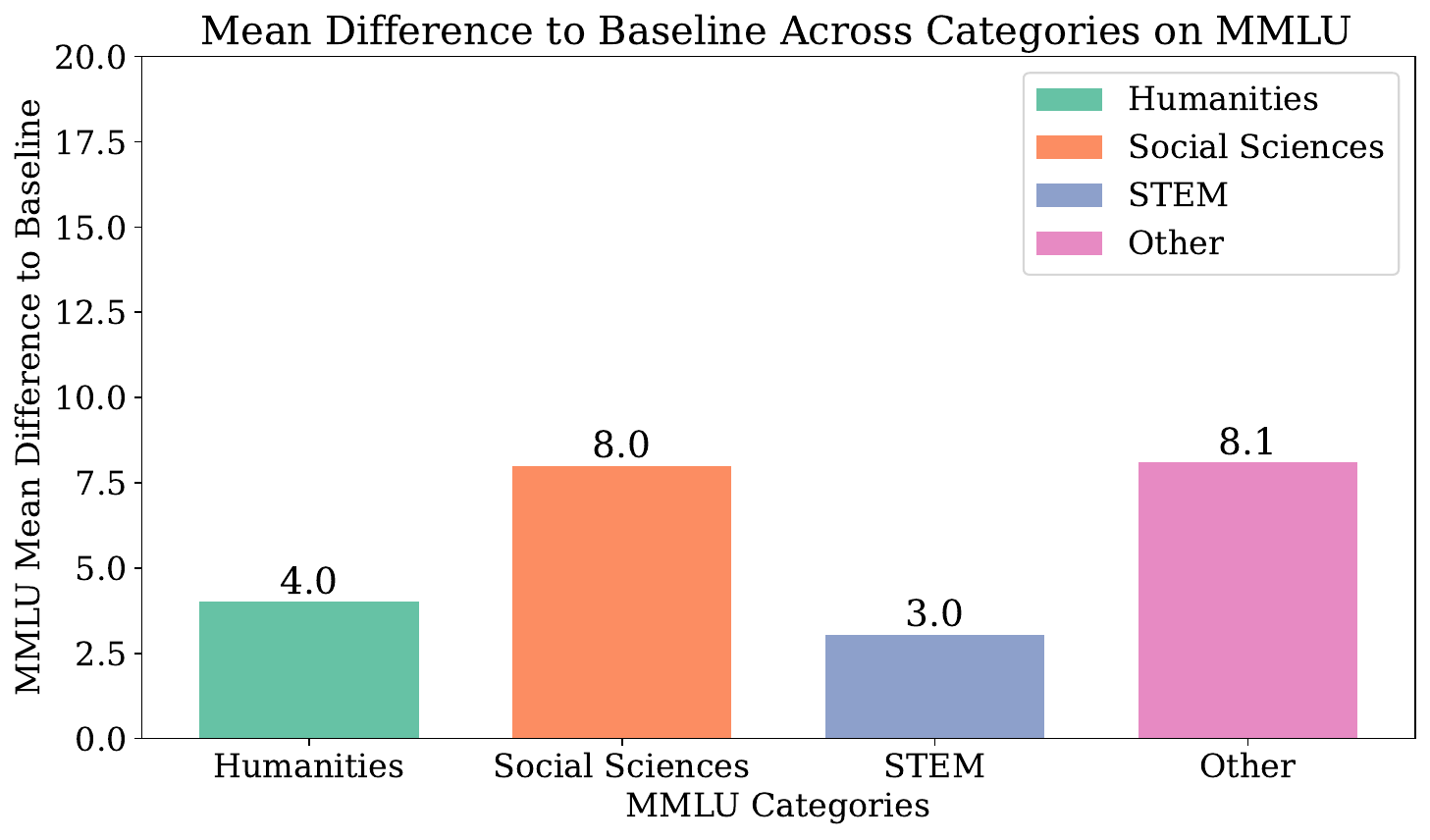}
    \caption{MMLU}
    \label{fig:difference_to_baseline_categories_mmlu}
\end{subfigure}
\vspace{-6mm}
\caption{Difference to the baseline for the XL model across categories on LegalBench and MMLU.}
\label{fig:difference_to_baseline_categories}
\vspace{-3mm}
\end{figure*}

To analyze the change in performance in more detail, we show the difference to the baseline for the XL model on LegalBench and MMLU across tasks (see \Cref{fig:difference_to_baseline_tasks}) and across categories (see \Cref{fig:difference_to_baseline_categories}). 
We find that FLawN-T5 outperforms baseline Flan-T5 in most LegalBench tasks in most categories. The exception are tasks in the interpretation category, specifically CUAD \citep{hendrycks_cuad_2021}, where the fine-tuned model is actually worse than the baseline by around 10 points on average. A possible explanation could be negative transfer from the instruction tuning data since the task formulations are very different to the instructions in LegalBench. In MAUD \citep{wang_maud_2023} and Contract-NLI \citep{koreeda_contractnli_2021}, the instructions are much more similar from LawInstruct to LegalBench, leading to improvements compared to the baseline.
On MMLU, most categories and tasks see increases in performance, especially the categories social sciences and other. We find that performance suffers mostly in the STEM category and to some extent in the humanities. Interestingly, the largest drop is in machine learning but the largest rise is in high school computer science. In the humanities, more ``hard'' disciplines are affected by performance decrease, such as formal logic and logical fallacies. 

Across categories overall we see lower improvements in conclusion and interpretation. Conclusion is one of LegalBench categories requiring more sophisticated reasoning capabilities; maybe larger models would see larger gains there. Concurrent work \citep{colombo_saullm-7b_2024} instruction tuned on synthetic legal data. They even saw a drop in performance in conclusion tasks compared to the baseline arguing, that conclusion tasks \say{require much more
pure deductive reasoning than actual legal knowledge} compared to tasks from the other categories.
Lower improvement in interpretation could be explained by negative transfer caused through different instructions in CUAD. Our hypothesis of a potential negative transfer is corroborated by our results on LegalBench by categories when we remove the datasets or tasks that overlap between LawInstruct and LegalBench (see \Cref{fig:difference_to_baseline_categories_lb_heldout}): We see larger gains compared to the baseline for both the conclusion and the interpretation categories.

\section{Ablations}
\label{sec:ablations}
In this section, we perform controlled experiments across the starting checkpoints, data mixtures, instruction styles and amount of instruction tuning data during pretraining. We show additional ablations regarding sampling styles, licenses and crosslingual transfer from multilingual data in \Cref{sec:additional_ablations}. Flan-T5 performs best in the studied parameter ranges. Baselines for other models are in \Cref{sec:detailed_results} \Cref{tab:paper_baselines}.

    \subsection{Starting Checkpoint}

    \begin{figure}[ht]
    \centering
    \includegraphics[width=0.95\columnwidth]{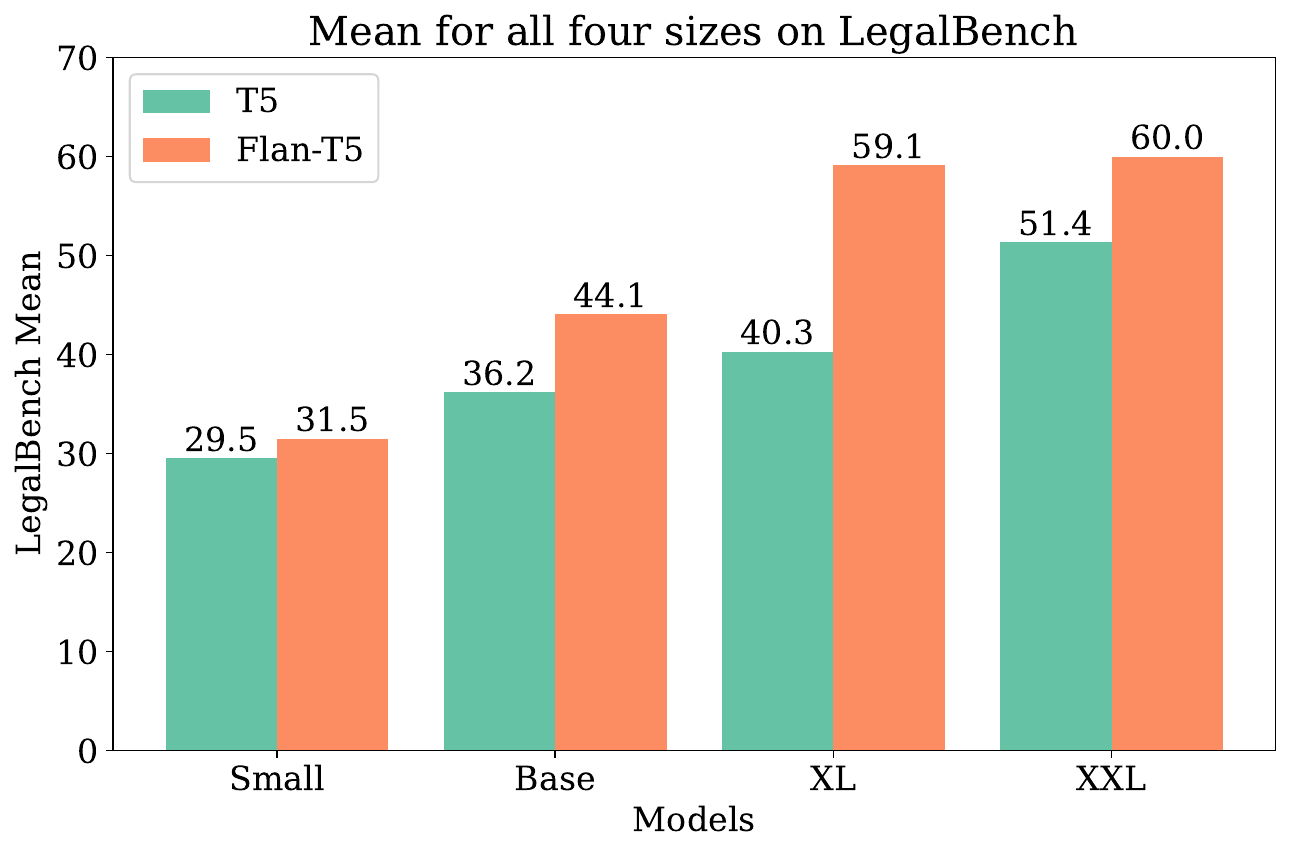}
    \vspace{-3mm}
    \caption{Starting instruction tuning from the Flan-T5 checkpoint improves results across all sizes.}
    \label{fig:starting_checkpoint_ift}
    \end{figure}

    \emph{Should you start in-domain instruction tuning from a base model or from an instruction tuned model?}
    $\Rightarrow$ \textbf{Starting from an instruction tuned model is better across sizes except Small.}
    In \Cref{fig:starting_checkpoint_ift}, we compare instruction tuning from a base T5 and a Flan-T5 model in four different sizes (Small, Base, XL and XXL) (detailed results in \Cref{sec:detailed_results} \Cref{tab:starting_checkpoint_ift}). We find that for the larger sizes, the instruction tuned Flan-T5 is a better starting point ($p < 0.001$), leading to higher performance on LegalBench. For the Small size the difference is not statistically significant ($p = 0.058$). 
    We use the Flan-T5 model as a starting point in all experiments unless specified otherwise.

    \subsection{Data Mixture}

    \begin{figure}[ht]
    \centering
    \includegraphics[width=0.95\columnwidth]{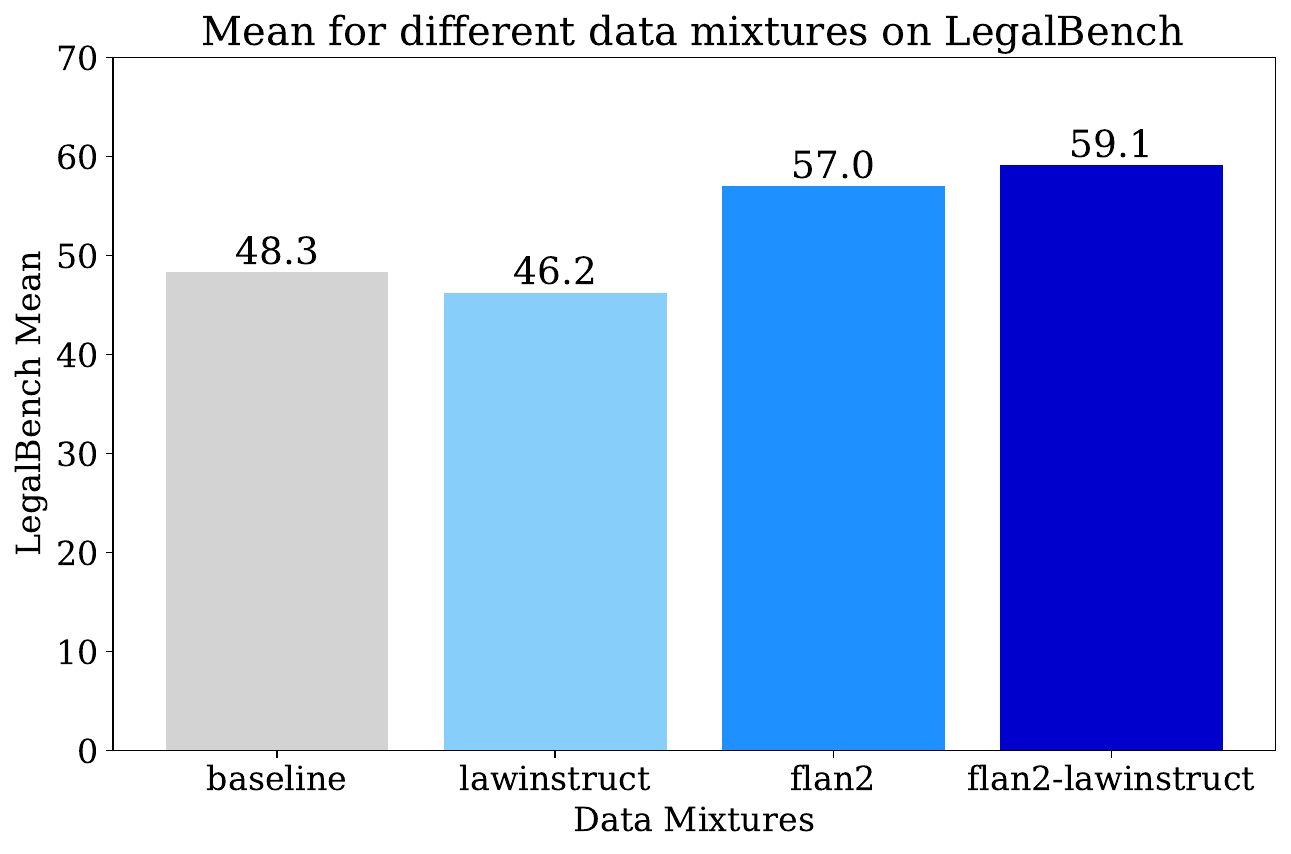}
    \vspace{-3mm}
    \caption{Accuracy of the Flan-T5 XL model on LegalBench using three data mixtures.}
    \label{fig:data_mixtures}
    \end{figure}
    
    \emph{What data mixtures should you choose for in-domain instruction tuning?}
    $\Rightarrow$ \textbf{Mixing in general instruction tuning datasets is necessary.}
    In \Cref{fig:data_mixtures}, we compare instruction tuning with three different data mixtures: lawinstruct, flan2 \citep{chung_scaling_2022}, and flan2-lawinstruct (where we sample equally from flan2 and lawinstruct) (detailed results in \Cref{sec:detailed_results} \Cref{tab:data_mixtures}). Interestingly, when only training on lawinstruct, downstream accuracy drops, possibly due to the instructions in our datasets being formulated differently than the original Flan instructions. Training on flan2 and flan2-lawinstruct leads to an aggregate increase of 7.7 points (48.3 to 56) and 10.8 points (48.3 to 59.1) respectively.
    We use the flan2-lawinstruct mixture in all experiments unless specified otherwise.


    \subsection{Instruction Style}

\begin{figure*}[h]
    \centering
    \begin{subfigure}[b]{0.49\textwidth}
        \centering
        \includegraphics[width=1\columnwidth]{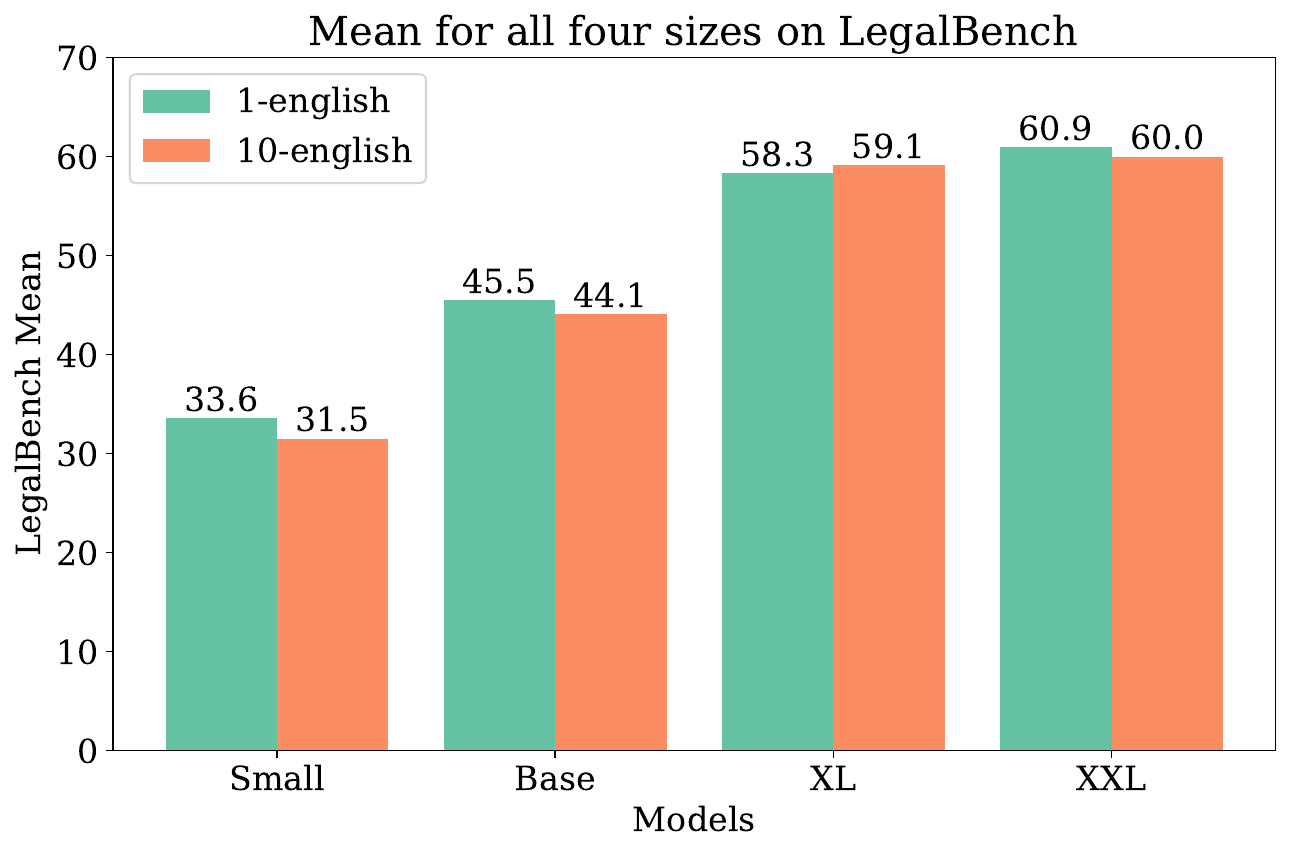}
        \caption{English}
        \label{fig:instruction_style_english}
    \end{subfigure}%
    \hfill 
    \begin{subfigure}[b]{0.49\textwidth}
        \centering
        \includegraphics[width=1\columnwidth]{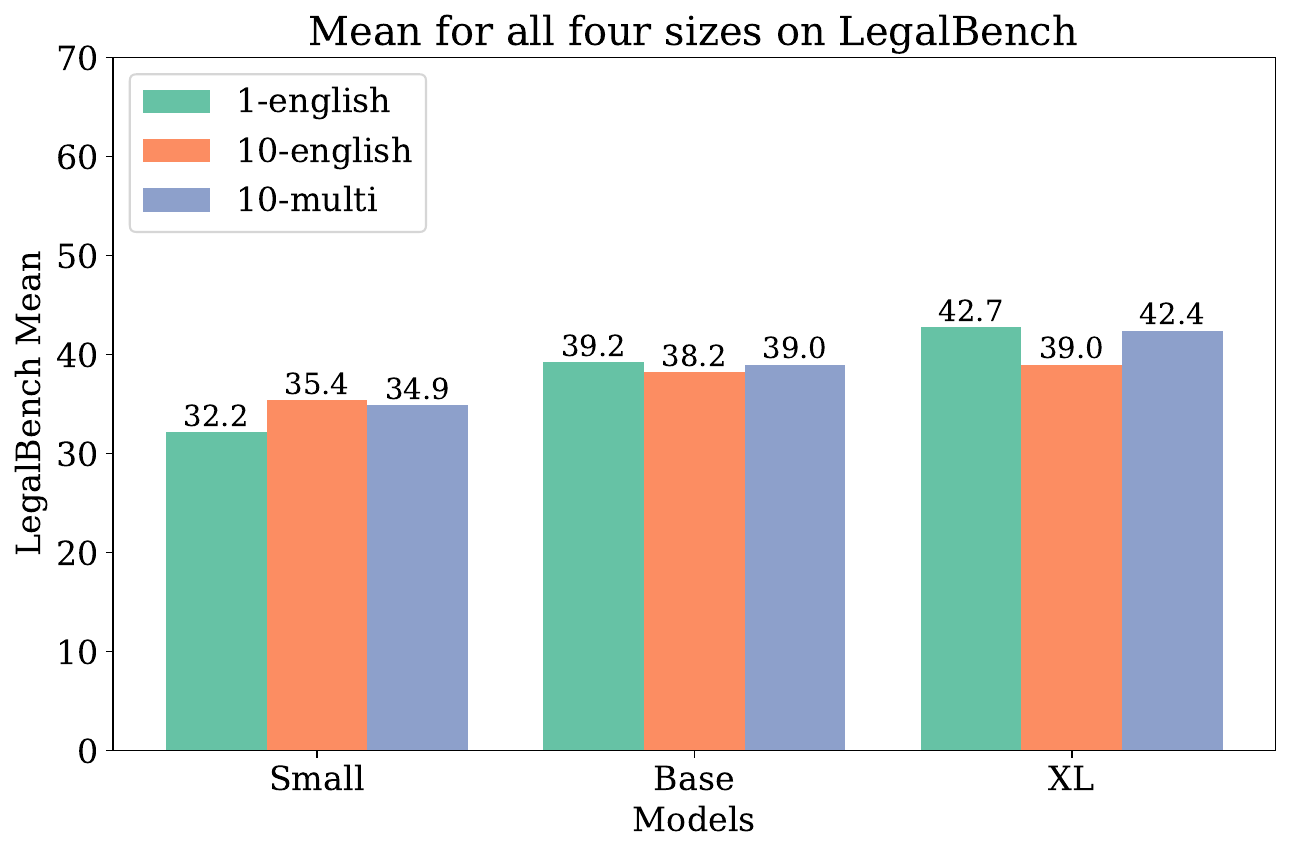}
        \caption{Multilingual}
        \label{fig:instruction_style_multi}
    \end{subfigure}
    \vspace{-3mm}
    \caption{Ablation on the instruction style on English/multilingual flan2-lawinstruct from the Flan-T5/mT5 checkpoint across all sizes.}
    \label{fig:instruction_style}
    \vspace{-3mm}
    \end{figure*}

    \emph{Are models trained with more diverse instructions better on LegalBench?}
    $\Rightarrow$ \textbf{Results are mixed, overall just using one instruction is probably sufficient.}
    In \Cref{fig:instruction_style}, we compare the performance of training with just one manually written instruction vs.\ ten paraphrased instructions with GPT-4 from one seed instruction, all else constant (detailed results in \Cref{sec:detailed_results} \Cref{tab:instruction_style_english}). For Flan-T5 (see \Cref{tab:instruction_style_english}), for Small, one instruction is better than ten ($p = 0.035$); for the other sizes we find no difference. For mT5 (see \Cref{fig:instruction_style_multi}), for Small, one instruction is worse than ten both monolingual ($p = 0.005$) and multilingual ($p = 0.01$) whereas for XL, ten English instructions underperform one English ($p < 0.001$) and ten multilingual ones ($p < 0.001$). In aggregate, differences are small without a consistent trend.

    \subsection{Amount of Instruction Data During Continued Pretraining}
    \emph{How much instruction tuning data should be mixed in during continued pretraining?}
    $\Rightarrow$ \textbf{Continued pretraining seems to be rather robust w.r.t. the amount of instruction tuning samples mixed in.}
    In \Cref{tab:domain_adaptation_small,tab:domain_adaptation_base,tab:domain_adaptation_xl,tab:domain_adaptation_xxl}, we investigate the benefit of mixing varying amounts of instruction tuning data in during continued pretraining (detailed results in \Cref{sec:detailed_results} \Cref{tab:domain_adaptation_small,tab:domain_adaptation_base,tab:domain_adaptation_xl,tab:domain_adaptation_xxl}). We compare results on LegalBench of instruction tuning runs after 10K to 90K steps of continued pretraining. For the Small model, the benefit of continued pretraining over just instruction tuning is significant (34.9 for just instruction tuning vs.\ ~40 after continued pretraining). Conversely, for the XL model, continued pretraining often underperforms compared to just instruction tuning. For the XXL model, more instruction tuning samples during continued pretraining improve performance, unlike for the Small and XL models. Across sizes, continued pretraining's effectiveness appears robust to the number of instruction tuning samples used.\footnote{Mixing instruction tuning data during continued pretraining without more instruction tuning does not improve results.}





\section{Related Work}

Domain-specific pretraining, covering areas such as medicine, law, and science, significantly enhances Language Model performance on related tasks \citep{Beltagy2019SciBERTAP, Gu2021, chalkidis-etal-2020-legal}. SciBERT \citep{Beltagy2019SciBERTAP}, for instance, was pretrained on a mix of computer science and biomedical papers, exemplifying this approach in the scientific domain. Other models like PubMedBERT \citep{Gu2021} and BioBERT \citep{Lee2020}, specifically pretrained on biomedical datasets, 
have shown improvements in medical NLP tasks \citep{Huang2019}.

\subsection{Domain-specific Legal Pretraining}

In the legal domain, models such as LegalBERT, pretrained on 12 GB of English legal texts, demonstrated notable success in domain-specific challenges \citep{chalkidis-etal-2020-legal}. CaseLaw-BERT capitalized on the English Harvard Law case corpus spanning from 1965 to 2021 \citep{zheng_pretrained_2022}, while \citet{niklaus_budgetlongformer_2022} pretrained LongFormer models on the Pile-of-Law \citep{henderson_pile_2022} using the replaced token detection task \citep{clark_electra_2020} for enhanced performance. Further advancements were made by \citet{chalkidis2023lexfiles}, who developed new English legal LMs yielding superior results on LexFiles, a compilation of 11 sub-corpora from six English-speaking legal systems encompassing 19B tokens. Additionally, \citet{niklaus_multilegalpile_2024} introduced a vast multilingual legal corpus, training both monolingual and multilingual legal models to achieve state-of-the-art results on LexGLUE \citep{LexGLUE} and LEXTREME \citep{niklaus_lextreme_2023}. Models have also been developed for specific jurisdictions, including the Swiss \citep{rasiah_scale_2023}, Italian \citep{licari_italian-legal-bert_2022}, Romanian \citep{masala-etal-2021-jurbert}, and Spanish \citep{gutierrez-fandino_spanish_2021} legal systems. Despite the prevalence of smaller encoder-based legal-specific LMs, larger generative models in this space remain scarce. This work seeks to bridge that gap.

\subsection{Instruction Tuning} 


Instruction tuning -- the process of finetuning auto-regressive pretrained language models on corpora of reciprocal instruction--response pairs -- has emerged as a critical step for building responsive models that are useful for many tasks \citep{instructgpt, chowdhery_palm_2022, wei_finetuned_2022, sanh_multitask_2022}. Some go as far as to claim that this training paradigm is the key to imbuing language models with the generalized capability of zero-shot instruction following behavior \citep{chung_scaling_2022}. 
Instruction tuning refers to few-shot or zero-shot adaptation of large language models to new tasks, where the task is described in natural language in the training examples. Following \citet{wei-etal-2021-finetuned}, it is common to transform existing datasets into instruction datasets by manually composing templates and filling these with specific examples. It is through these domain-specific training procedures that we build and evaluate legal data adaptation in LLMs.

\section{Conclusion}
We curated LawInstruct, the first instruction tuning dataset for the legal domain by aggregating various high-quality annotated datasets and writing instructions for the different tasks. We used LawInstruct to instruction tune T5 based models, creating FLawN-T5 and a new state-of-the-art on LegalBench in all investigated parameter sizes. We openly release LawInstruct on Hugging Face.

\section*{Limitations \& Future Work}

Our use of template-based instruction creation may restrict the variety of instructions, potentially affecting the model's ability to handle more diverse or novel legal queries effectively. While we already tried to address this by paraphrasing the instructions with GPT-4, the diversity may still be limited. 
To alleviate this problem, we could create synthetic data either by generating responses from instructions \citep{wang-etal-2023-self-instruct} or reversely, by generating instructions to responses \citep{koksal_longform_2024}. It is important to take care to do detailed quality checks since hallucinated content may hurt more than improve, especially in the legal domain.
Another way to alleviate this diversity problem is working with  legal professionals to identify and annotate new tasks for the legal domain. However, this route is out of reach for many academic efforts due to large salaries of qualified lawyers.

It would be great to conduct a qualitative human expert study on the model outputs. Unfortunately, a rigorous human evaluation with qualified legal experts across multiple jurisdictions was beyond our current resources. As LegalBench and MMLU are multiple-choice QA benchmarks, the utility of human evaluation is limited compared to long-form generative tasks. The objective nature of these benchmarks allows for reliable automated evaluation. However, as future work expands to evaluating on more open ended generation tasks we definitely recommend conducting human evaluations.

To our surprise, continued pretraining only benefited at the Small model size, but not at larger sizes. Due to our focus on instruction tuning and limited budget, we were not able to study this effect in more detail. In future work, we would like to study the robustness of our findings across model sizes. We hypothesize that methods like mixing in data from the original training set, using smaller learning rates, and adding loss terms to discourage the weights to depart too much from the original model could potentially lead to different conclusions.


In the future, we would like to extend LawInstruct with more high-quality datasets released after our experiments such as 
Long-form Legal Question Answering \citep{louis_interpretable_2023}, 
Keyphrase Generation \citep{salaun_europa_2024}, 
Negation Scope Resolution \citep{christen_resolving_2023}, 
or Legal Violation Detection \citep{bernsohn_legallens_2024}. 

Additionally, it would be interesting to investigate the overlap between the T5 pretraining dataset C4 and the MultiLegalPile to get a better understanding of the potential benefits of continued pretraining.



\section*{Acknowledgements}
We would like to thank Baq Haidri and Irhum Shafkat for help with debugging and John Kirchenbauer, Sang Michael Xie and David Hall for feedback on the project. Thanks to Neel Guha for help with debugging LegalBench and for feedback on analyzing results. Thanks to the Stanford NLP group and the anonymous reviewers for feedback on the draft.

\bibliography{anthology,references,custom}

\appendix

\section{Use of AI Assistants}
We used ChatGPT 3.5 and 4 for shortening texts and editing LaTeX more efficiently.

\section{Detailed Dataset Description}
\label{sec:detailed_dataset_description}

\begin{figure*}[ht]
\centering
\includegraphics[width=\textwidth]{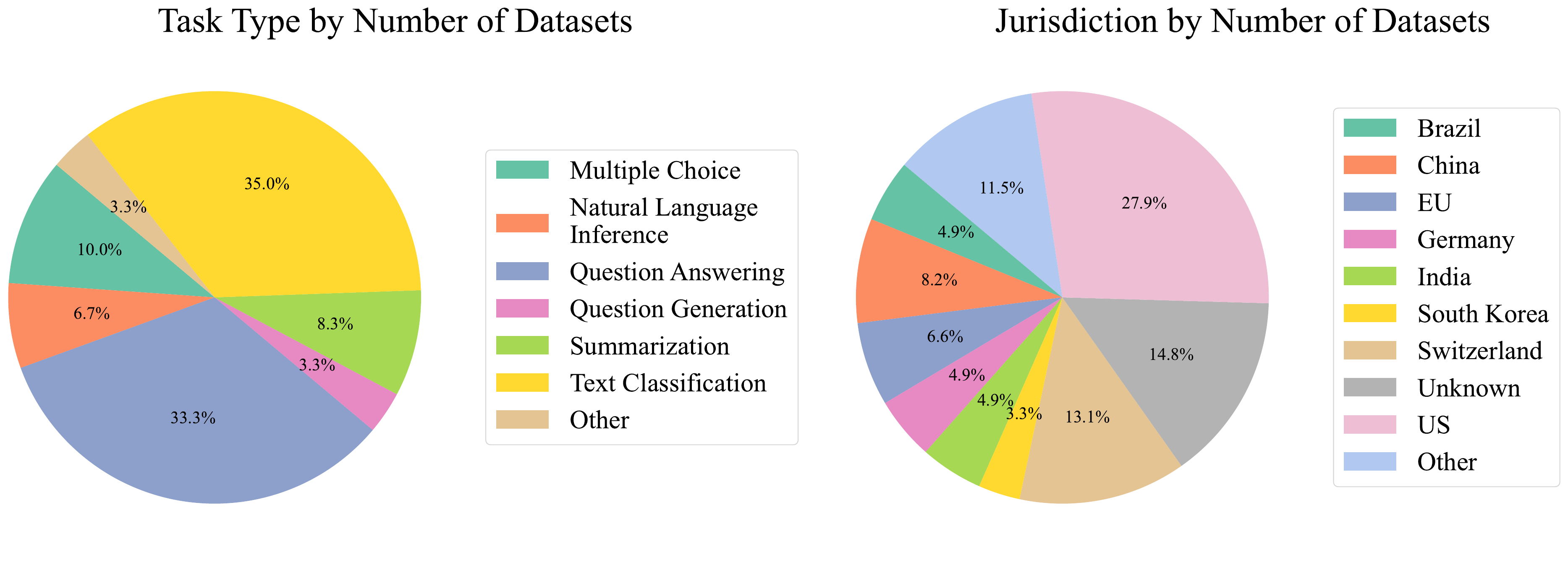}
\caption{Jurisdiction and task type by datasets.}
\label{fig:lawinstruct_by_datasets}
\end{figure*}

\Cref{fig:lawinstruct_by_datasets} shows the LawInstruct task type and jurisdiction composition by dataset. \Cref{tab:dataset} lists the dataset (and sources), license, language, jurisdiction, task type, subtask, and number of examples for each dataset included in LawInstruct.

\begin{table*}[ht!p]
    \centering
    \caption{Overview of the LawInstruct datasets. The 24 EU langs are \lang{bg},  \lang{cs}, \lang{da}, \lang{de}, \lang{el}, \lang{en}, \lang{es}, \lang{et}, \lang{fi}, \lang{fr}, \lang{ga}, \lang{hu}, \lang{it}, \lang{lt}, \lang{lv}, \lang{mt}, \lang{nl}, \lang{pt}, \lang{ro}, \lang{sv}, \lang{sk}. Abbreviations: Terms of Service (ToS)}
    \begin{adjustbox}{max width=1.04\linewidth,center} 
    \begin{tabular}{@{} p{11.2cm} l p{2.7cm} l l l r @{}}
    \toprule
    Dataset & License & Languages & Jurisdiction & Tasks & Subtask & Examples \\
    \midrule
    Benchmark for Understanding Indian Legal Documents (BUILD) \citep{kalamkar-etal-2022-named} & Unknown & \lang{en} & India & Text classification & Rhetorical role & 28,986 \\
    Brazilian Bar Exam \citep{delfino2017passing} & Unknown & \lang{pt} & Brazil & Question answering & Bar exam questions & 2,130 \\
    Brazilian Court Decisions \citep{lage-freitas_predicting_2022} & Unknown & \lang{pt} & Brazil & Text classification & Judgment & 3,234\\
    Brazilian Court Decisions \citep{lage-freitas_predicting_2022} & Unknown & \lang{pt} & Brazil & Text classification & Decision Unanimity & 1,715\\
    BrCAD5 \citep{jacob_de_menezes-neto_using_2022} & CC BY-NC-SA 4.0 & \lang{pt} & Brazil & Multiple choice & Judgment & 1,225,922 \\
    BrCAD5 \citep{jacob_de_menezes-neto_using_2022} & CC BY-NC-SA 4.0 & \lang{pt} & Brazil & Text classification & Judgment & 612,961  \\
    BrCAD5 \citep{jacob_de_menezes-neto_using_2022} & CC BY-NC-SA 4.0 & \lang{pt} & Brazil & Text classification & Area of law & 612,961  \\
    BrCAD5 \citep{jacob_de_menezes-neto_using_2022} & CC BY-NC-SA 4.0 & \lang{pt} & Brazil & Text classification & Topic & 1,838,883  \\
    BVADecisions \citep{walker2019automatic} & MIT & \lang{en} & USA & Text classification & Rhetorical role & 8,818 \\
    BVADecisions \citep{walker2019automatic} & MIT & \lang{en} & USA & Question answering & Relevant rules & 2 \\
    CAIL 2019 \citep{xiao2019cail2019} & Unknown & \lang{zh} & China & Question answering & Chinese legal case questions & 39,333 \\
    CAIL 2022 \citep{cail2022} & Unknown  & \lang{zh} & China & Text classification & Charge/crime & 10,448 \\
    CAIL 2022 \citep{cail2022} & Unknown  & \lang{zh} & China & Argument \& counter-argument & 5,224 \\
    CAIL 2022 \citep{cail2022} & Unknown & \lang{zh} & China & Question answering & Response to argument & 5,224 \\
    Case Briefs \citep{casebriefs2024} & CC BY-NC & \lang{en} & USA & Question answering & Legal analysis of facts & 2,619 \\
    CaseHOLD \citep{zheng-casehold} & CC-BY & \lang{en} & USA & Multiple choice & Legal holding statements & 45,000 \\
    Change My View \citep{tan+etal:16a} & Unknown & \lang{en} & N/A & Argument \& counter-argument & 3,456 \\
    COLIEE \citep{kim2022coliee} & Academic use only & \lang{en}, \lang{jp} & Canada/Japan & Question generation & Entailed question & 1,774 \\
    COLIEE \citep{kim2022coliee} & Academic use only  & \lang{en}, \lang{jp} & Canada/Japan & Natural language inference & Passage entailment & 125,954 \\
    COLIEE \citep{kim2022coliee} & Academic use only  & \lang{en}, \lang{jp} & Canada/Japan & Question answering & Relevant legal rule & 1,774 \\
    ContractNLI \citep{koreeda_contractnli_2021} & CC BY-NC & \lang{en} & USA & Natural language inference & Premise hypothesis entailment & 14,010 \\
    COVID-19 Emergency Measures (EXCEPTIUS) \citep{tziafas2021multilingual} & Unknown & \lang{en}, \lang{fr}, \lang{hu}, \lang{it}, \lang{nb}, \lang{nl}, \lang{pl} & EU & Text classification & Measure type & 3,312 \\
    European Court of Human Rights (ECtHR) \citep{chalkidis_paragraph-level_2021} & CC BY-NC-SA 4.0 & \lang{en} & EU & Text classification (multi-label) & Violated article & 9,000 \\
    European Court of Human Rights (ECtHR) \citep{chalkidis_paragraph-level_2021} & CC BY-NC-SA 4.0 & \lang{en} & EU & Text classification (multi-label) & Allegedly violated article & 9,000 \\
    EOIR \citep{henderson_pile_2022} & CC BY-NC-SA 4.0 & \lang{en} & USA & Text classification & Pseudonymity & 8,089 \\
    EURLEX \citep{chalkidis_large-scale_2019} & CC BY-SA 4.0 & \lang{en} & EU & Text classification & EuroVoc core concepts & 55,000 \\
    EUR-Lex-Sum \citep{aumiller_eur-lex-sum_2022} & CC BY 4.0 & 24 EU langs & EU & Summarization & EU Legal Acts & 22,989 \\
    German Argument Mining \citep{urchs_design_2021} & CC BY 4.0 & \lang{de} & Germany & Text classification & Argumentative function & 19,271 \\
    German Rental Agreements \citep{steinberger-etal-2006-jrc} & Unknown & \lang{de} & Germany & Text classification & Semantic type & 3,292 \\
    Greek Legal Code \citep{papaloukas-etal-2021-multi} & CC BY 4.0 & \lang{el} & Greece & Text classification & Volume (coarse thematic topic) & 28,536 \\
    Greek Legal Code \citep{papaloukas-etal-2021-multi} & CC BY 4.0 & \lang{el} & Greece & Text classification & Chapter (intermediate thematic topic) & 28,536 \\
    Greek Legal Code \citep{papaloukas-etal-2021-multi} & CC BY 4.0 & \lang{el} & Greece & Text classification & Subject (fine-grain thematic topic) & 28,536 \\
    Greek Legal NER (elNER) \citep{bartziokas2020datasets} & CC BY-NC-SA 4.0 & \lang{el} & Greece & Named entity recognition & Greek legal entities & 17,699 \\
    ILDC \citep{malik_ildc_2021} & CC BY-NC & \lang{en} & India & Text classification & Judgment & 37,387 \\
    International Citizenship Law \citep{vink2021globalcit} & CC BY 4.0 & \lang{en} & International & Question answering & Citizenship acquisition & 6,460 \\
    International Citizenship Law \citep{vink2021globalcit} & CC BY 4.0 & \lang{en} & International & Question answering & Citizenship loss & 2,850 \\
    JEC-QA \citep{zhong2020jec} & CC BY-NC-ND & \lang{zh} & China & Multiple choice & National Judicial Examination of China & 21,072 \\
    Korean Legal QA \citep{heewon2021} & Academic use only & \lang{ko} & South Korea & Question answering & Relevant law & 1,830 \\
    LawngNLI \citep{bruno_lawngnli_2022} & MIT & \lang{en} & USA & Natural language inference & Premise hypothesis entailment & 1,142,304 \\
    LBOX OPEN \citep{hwang_multi-task_2022} & CC BY-NC & \lang{ko} & South Korea & Text classification & Judgment & 12,142 \\
    LBOX OPEN \citep{hwang_multi-task_2022} & CC BY-NC & \lang{ko} & South Korea & Text classification & Relevant statutes & 13,317 \\
    LEDGAR \citep{tuggener_ledgar_2020} & CC BY-NC & \lang{en} & USA & Text classification & Contract provision category & 60,000 \\
    Legal Case Document Summarization \citep{shukla_legal_2022, bhattacharya2019comparative} & CC BY-SA & \lang{en} & India & Summarization & Indian Supreme Court & 7,080 \\
    Legal Case Summarization \citep{shukla_legal_2022, bhattacharya2019comparative} & CC BY-SA & \lang{en} & UK & Summarization & UK Supreme Court & 693 \\
    LegalNERo \citep{pais_vasile_2021_4922385} & CC0 1.0 & \lang{ro} & Romania & Named entity recognition & Romanian legal entities & 7,552 \\
    LegalQA \citep{siatnlp2019} & Unknown & \lang{zh} & China & Question answering & Legal advice & 21,946 \\
    LeNER-Br \citep{luz_de_araujo_lener-br_2018} & Unknown & \lang{pt} & Brazil & Named entity recognition & Brazilian legal entities & 7,828 \\
    Littleton \citep{basu2022programming} & MIT & \lang{en} & USA & Question answering & Relevant future interests & 131 \\
    Littleton \citep{basu2022programming} & MIT & \lang{en} & USA & Question answering & Event graph & 143 \\
    MAPA \citep{de-gibert-bonet-etal-2022-spanish} & CC BY-NC 4.0 & 24 EU langs & EU & Named entity recognition & Coarse-grained & 27,823 \\
    MAPA \citep{de-gibert-bonet-etal-2022-spanish} & CC BY-NC 4.0 & 24 EU langs & EU & Named entity recognition & Fine-grained & 27,823 \\
    MAUD \citep{wang_maud_2023} & CC BY & \lang{en} & USA & Multiple choice & Merger agreement questions & 10,751 \\ 
    MAUD \citep{wang_maud_2023} & CC BY & \lang{en} & USA & Text classification & Deal point category & 25,827 \\ 
    MAUD \citep{wang_maud_2023} & CC BY & \lang{en} & USA & Text classification & Question type & 25,827 \\ 
    MAUD \citep{wang_maud_2023} & CC BY & \lang{en} & USA & Text classification & Text type & 25,827 \\ 
    Mining Legal Arguments \citep{Habernal.et.al.2022.arg} & Apache-2.0 & \lang{en} & EU & Named entity recognition & Actors & 31,852 \\
    Mining Legal Arguments \citep{Habernal.et.al.2022.arg} & Apache-2.0 & \lang{en} & EU & Named entity recognition & Argument type & 31,852 \\
    MultiEURLEX \citep{chalkidis_multieurlex_2021} & CC BY-SA & 24 EU langs & EU & Text classification (multi-label) & EuroVoc taxonomy (coarse level) & 1,265,000 \\
    MultiEURLEX \citep{chalkidis_multieurlex_2021} & CC BY-SA & 24 EU langs & EU & Text classification (multi-label) & EuroVoc taxonomy (intermediate level) & 911,798 \\
    MultiEURLEX \citep{chalkidis_multieurlex_2021} & CC BY-SA & 24 EU langs & EU & Text classification (multi-label) & EuroVoc taxonomy (fine-grain level) & 1,265,000 \\
    Multi-LexSum \citep{shen_multi-lexsum_2022} & ODC-By & \lang{en} & USA & Summarization & Long to short & 2,210 \\
    Multi-LexSum \citep{shen_multi-lexsum_2022} & ODC-By & \lang{en} & USA & Summarization & Long to tiny & 1,130 \\
    Multi-LexSum \citep{shen_multi-lexsum_2022} & ODC-By & \lang{en} & USA & Summarization & Short to tiny & 1,129 \\
    Natural Instructions (BillSum) \citep{kornilova_billsum_2019} & CC0 1.0 & \lang{en} & USA & Summarization & U.S Congressional and California state bills &  25,200 \\
    Natural Instructions (CAIL 2018) \citep{xiao_cail2018_2018} & Unknown & \lang{zh} & China & Question answering & Judgment & 5,988 \\
    Natural Instructions (CaseHOLD) \citep{zheng-casehold} & CC-BY & \lang{en} & USA & Multiple choice & Correct answer & 5,988 \\
    Natural Instructions (CaseHOLD) \citep{zheng-casehold} & CC-BY & \lang{en} & USA & Multiple choice & Incorrect answer & 5,988 \\
    Natural Instructions (CUAD) \citep{hendrycks_cuad_2021} & CC BY 4.0 & \lang{en} & Question answering &  Information relevant for contract review & 2,442 \\
    Natural Instructions (CUAD) \citep{hendrycks_cuad_2021} & CC BY 4.0 & \lang{en} & USA & Question generation & Questions relevant for contract review & 2,442 \\
    Natural Instructions (EURLEX) \citep{chalkidis_large-scale_2019} & CC BY-SA 4.0 & \lang{en} & EU & Text classification & Regulation, decisions, or directive & 5,850 \\
    Natural Instructions (EURLEX) \citep{aumiller_eur-lex-sum_2022} & CC BY-SA 4.0 & \lang{en} & EU & Summarization & EU Legal Acts & 3,900 \\
    Natural Instructions (OPP-115) \citep{wilson2016creation} & CC BY-NC & \lang{en} & USA & Question answering & Type of information used by website & 18,480 \\
    Natural Instructions (OPP-115) \citep{wilson2016creation} & CC BY-NC & \lang{en} & USA & Question answering & Purpose of privacy policy & 18,474 \\
    Natural Instructions (Overruling) \citep{zheng-casehold} & Unknown & \lang{en} & USA & Text classification & Sentence is overruling & 14,370 \\
    OLC Memos \citep{henderson_pile_2022} & CC BY-NC & \lang{en} & USA & Question answering & Write a legal research memo & 1,038 \\
    Online ToS \citep{drawzeski_corpus_2021} & CC BY-NC 2.5 & \lang{de}, \lang{en}, \lang{it}, \lang{pt} & Unknown & Text classification & Clause topic & 19,942 \\
    Online ToS \citep{drawzeski_corpus_2021} & CC BY-NC 2.5 & \lang{de}, \lang{en}, \lang{it}, \lang{pt} & Unknown & Text classification & Unfair contractual term type & 2,074 \\
    Plain English Contracts Summarization \citep{manor-li-2019-plain} & Unknown & \lang{en} & USA & Summarization & Software licenses, ToS & 446 \\
    PrivacyQA \citep{ravichander-etal-2019-question} & MIT & \lang{en} & Unknown & Question answering & Contents of privacy policies & 185,200 \\
    PrivacySummarization \citep{keymanesh2020toward} & MIT & \lang{en} & USA & Summarization & Privacy policies, ToS, and cookie policies & 5,751 \\
    RedditLegalQA \citep{henderson_pile_2022} & CC BY 4.0 & \lang{en} & Unknown & Question answering & Legal advice from r/legaladvice & 192,953 \\
    Sara \citep{holzenberger2020statutory} & Unknown & \lang{en} & USA & Natural language entailment & Fact entailment & 176 \\
    Sara \citep{holzenberger2020statutory} & Unknown & \lang{en} & USA & Question answering & Tax liability & 160 \\
    SaraProlog \citep{holzenberger2020statutory} & Unknown & \lang{en} & USA & Question answering & Fact pattern to prolog code & 376 \\
    SaraProlog \citep{holzenberger2020statutory} & Unknown & \lang{en} & USA & Question answering & Tax statute to prolog code & 9 \\
    Short Answer Feedback \citep{filighera-etal-2022-answer} & CC BY 4.0 & \lang{de} & Germany & Question answering & Answer question about German law & 1,596 \\
    Short Answer Feedback \citep{filighera-etal-2022-answer} & CC BY 4.0  & \lang{de} & Germany & Question answering & Feedback rating for answer & 1,596 \\
    Spanish Labor Law \citep{calleja2021bilingual} & CC BY 4.0 & \lang{es} & Spain & Extractive question answering & Answer question about Spanish labor law & 111 \\
    StackExchange Questions (Law) \citep{stackexchange2024} & CC BY-SA & \lang{en} & Unknown & Question answering & Online legal forum & 10,158 \\
    The Supreme Court Database \citep{Spaeth2020} & CC BY-NC 3.0 & \lang{en} & USA & Text classification & Issue areas & 5,000 \\
    Swiss Federal Supreme Court \citep{rasiah_scale_2023} & CC BY 4.0 & \lang{de}, \lang{fr} & Text generation & Case considerations sections (lower court) & 26 \\
    Swiss Courts \citep{rasiah_scale_2023} & CC BY 4.0 & \lang{de}, \lang{fr}, \lang{it} & Switzerland & Text generation & Case considerations sections (same court) & 234,313 \\
    Swiss Federal Supreme Court \citep{rasiah_scale_2023,stern_breaking_2024} & CC BY 4.0 & \lang{de}, \lang{fr}, \lang{it} & Switzerland & Text classification & Case criticality (based on citations) & 91,075 \\
    Swiss Courts \citep{rasiah_scale_2023,niklaus_swiss-judgment-prediction_2021} & CC BY 4.0 & \lang{de}, \lang{fr}, \lang{it}, \lang{en} & Switzerland & Multiple choice & Judgment & 477,636 \\
    Swiss Courts \citep{rasiah_scale_2023,niklaus_swiss-judgment-prediction_2021} & CC BY 4.0  & \lang{de}, \lang{fr}, \lang{it} & Switzerland & Text classification & Judgment & 385,719 \\
    Swiss Courts \citep{rasiah_scale_2023,niklaus_swiss-judgment-prediction_2021} & CC BY 4.0 & \lang{de}, \lang{fr}, \lang{it}, \lang{en} & Switzerland & Text classification & Area of law & 18,162 \\
    Swiss Courts \citep{rasiah_scale_2023,niklaus_swiss-judgment-prediction_2021} & CC BY 4.0  & \lang{de}, \lang{fr}, \lang{it}, \lang{en} & Switzerland & Text classification & Subarea of law & 18,162 \\
    Swiss Federal Supreme Court (Leading Decisions) \citep{rasiah_scale_2023} & CC BY 4.0  & \lang{de}, \lang{en}, \lang{fr}, \lang{it} & Switzerland & Text classification & Location (canton, region) & 42,342 \\
    Swiss Legislation \citep{rasiah_scale_2023} & CC BY 4.0 & \lang{de}, \lang{fr}, \lang{it}, \lang{rm} & Switzerland & Text classification & Abbreviation & 11,045 \\
    Swiss Legislation \citep{rasiah_scale_2023} & CC BY 4.0  & \lang{de}, \lang{en}, \lang{fr}, \lang{it}, \lang{rm} & Switzerland & Text classification & Canton & 35,698 \\
    Swiss Legislation \citep{rasiah_scale_2023} & CC BY 4.0 & \lang{de}, \lang{en}, \lang{fr}, \lang{it}, \lang{rm} & Switzerland & Text classification & Short description & 3,747 \\
    Swiss Legislation \citep{rasiah_scale_2023} & CC BY 4.0 & \lang{de}, \lang{en}, \lang{fr}, \lang{it}, \lang{rm} & Switzerland & Text classification & Title & 35,359 \\
    Thai Supreme Court Cases (TSCC)\citep{thanh2021alqac} & Academic use only & \lang{th} & Thailand & Question answering & Relevant legal articles (Thai Criminal Code) & 2,883 \\
    Turkish Constitutional Court \citep{mumcuoglu21natural} & CC BY 4.0 & \lang{tr} & Turkey & Multiple choice & Judgment & 1,804 \\
    Turkish Constitutional Court \citep{mumcuoglu21natural} & CC BY 4.0 & \lang{tr} & Turkey & Text classification & Judgment & 902 \\
    Unfair ToS \citep{Lippi2019} & Unknown & \lang{en} & USA & Text classification (multi-label) & Unfair contractual term type & 5,532 \\
    U.S Class Actions \citep{semo_classactionprediction_2022} & GPL-3.0 & \lang{en} & USA & Text classification & Judgment & 3,000 \\
    Valid Wills \citep{kwak-etal-2022-validity} & Unknown & \lang{en} & USA & Text classification & Statement supported by law/condition & 1,512 \\
    \bottomrule
    \end{tabular}
    \end{adjustbox}
    \label{tab:dataset}
\end{table*}
\section{Detailed Experimental Setup}
\label{sec:detailed_experimental_setup}

\subsection{Inexplicable Behaviour at the XXL Size}
\label{sec:inexplicable_xxl}

We spent considerable effort, including joint debugging with the authors of LegalBench, to reproduce their results. We double checked that the prompts, decoding hyperparameters and general setup are consistent. We conjecture, that the conversion of the Flan-T5 weights as done by Hugging Face on their hub leads to different behavior when running the models with T5X on TPUs (our setup) vs running them with Hugging Face transformers and PyTorch on NVIDIA GPUs (original LegalBench setup)\footnote{Similar issues are mentioned in this issue: \url{https://github.com/PiotrNawrot/nanoT5/issues/25}}. 

The XXL mT5 model did not train stably in the continued pretraining phase despite heavy hyperparameter tuning.

\subsection{Evaluation}
\label{sec:detailed_evaluation}

We excluded any legal tasks occurring in MMLU from LawInstruct. However, there is some overlap regarding the tasks included in LawInstruct and in LegalBench because high-quality legal tasks are rare. To control for these overlapping tasks, we evaluate on two versions of LegalBench holding out tasks by the datasets or tasks occurring in LawInstruct respectively.

\subsubsection{LegalBench Dataset Held Out}

If the source dataset of the LegalBench task occurs in LawInstruct, we remove it from the evaluation. Below, we list which tasks are overlapping. Overall 100 tasks are held out (see \Cref{tab:legalbench_dataset_held_out} for the complete list), so 61 tasks are remaining for LegalBench evaluation.

\begin{table*}
\centering
\caption{LegalBench Dataset Held Out}
\label{tab:legalbench_dataset_held_out}
\resizebox{\textwidth}{!}{
\begin{tabular}{p{3cm}p{17cm}p{3cm}}
\toprule
\textbf{Dataset} & \textbf{LawInstruct} & \textbf{LegalBench} \\
\midrule
\textbf{ContractNLI} & ContractNLI-contract\_nli & contract\_nli\_* \\
\midrule
\textbf{CUAD} & NaturalInstructionsLegal-cuad\_answer\_generation, NaturalInstructionsLegal-cuad\_question\_generation & cuad\_* \\
\midrule
\textbf{GLOBALCIT Citizenship Law Dataset} & InternationalCitizenshipLawQuestions-international\_citizenship\_law\_questions\_mode\_acq, InternationalCitizenshipLawQuestions-international\_citizenship\_law\_questions\_mode\_loss & international\_citizenship\_questions \\
\midrule
\textbf{MAUD} & MAUD-answer, MAUD-category, MAUD-question, MAUD-text\_type & maud\_* \\
\midrule
\textbf{OPP-115 (Online Privacy Policies, set of 115) Corpus} & NaturalInstructionsLegal-online\_privacy\_policy\_text\_information\_type\_generation, NaturalInstructionsLegal-online\_privacy\_policy\_text\_purpose\_answer\_generation & opp\_115\_* \\
\midrule
\textbf{Overruling} & NaturalInstructionsLegal-overruling\_legal\_classification & overruling \\
\midrule
\textbf{PrivacyQA} & PrivacyQA-privacy\_qa & privacy\_policy\_qa \\
\multicolumn{3}{c}{\textit{Note: The LegalBench privacy\_policy\_entailment Source field is currently incorrectly linked to this dataset (PrivacyQA), but is derived from a different dataset (APP-350 Corpus).}}\\
\midrule
\textbf{StAtutory Reasoning Assessment (SARA)} & Sara-sara\_entailment, Sara-sara\_tax\_liability, SaraProlog-sara\_prolog\_facts, SaraProlog-sara\_prolog\_statute & sara\_* (built off of SARA v2) \\
\midrule
\textbf{Unfair Terms of Service} & LexGLUE-unfair\_tos, LEXTREME-online\_terms\_of\_service\_clause\_topics (multilingual version), LEXTREME-online\_terms\_of\_service\_unfairness\_levels (multilingual version) & unfair\_tos \\
\bottomrule
\end{tabular}
}
\end{table*}

\subsubsection{LegalBench Task Held Out}

We additionally catalog instructions which train the LLM for a task captured in LegalBench. It is not necessary that the instruction-response pair in LawInstruct contain data from LegalBench, just that they are about similar legal tasks (e.g., classifying choice-of-forum provisions). In \Cref{tab:legalbench_task_held_out}, we list which tasks are overlapping. Overall 64 tasks are held out, so 97 tasks are remaining for LegalBench evaluation.


\begin{table*}[!ht]
\centering
\caption{LegalBench Task Held Out}
\label{tab:legalbench_task_held_out}
\resizebox{\textwidth}{!}{%
\begin{tabular}{p{3cm}p{7cm}p{7cm}}
\toprule
\textbf{Task} & \textbf{LawInstruct} & \textbf{LegalBench} \\
\midrule
\textbf{Rhetorical Role Labeling} & 
  bva\_decisions\_label, indian\_text\_segmentation, german\_argument\_mining & 
  function\_of\_decision\_section, oral\_argument\_question\_purpose \\
\midrule
\textbf{Civil Procedure Questions} & 
  civipro\_questions\_generate\_* & 
  diversity\_*, personal\_jurisdiction \\
\midrule
\textbf{Legal Entailment} & 
  coliee\_task3\_passage\_entailment, contract\_nli, lawng\_nli\_entailment & 
  contract\_nli\_* \\
\midrule
\textbf{Contractual Clause Classification} & 
  unfair\_tos, german\_rental\_agreements & 
  cuad\_*, jcrew\_blocker, unfair\_tos, contract\_qa \\
\bottomrule
\end{tabular}%
}
\end{table*}

\section{Additional Ablations}
\label{sec:additional_ablations}

    \subsection{Sampling Style}

    \begin{figure}[ht]
    \centering
    \includegraphics[width=0.95\columnwidth]{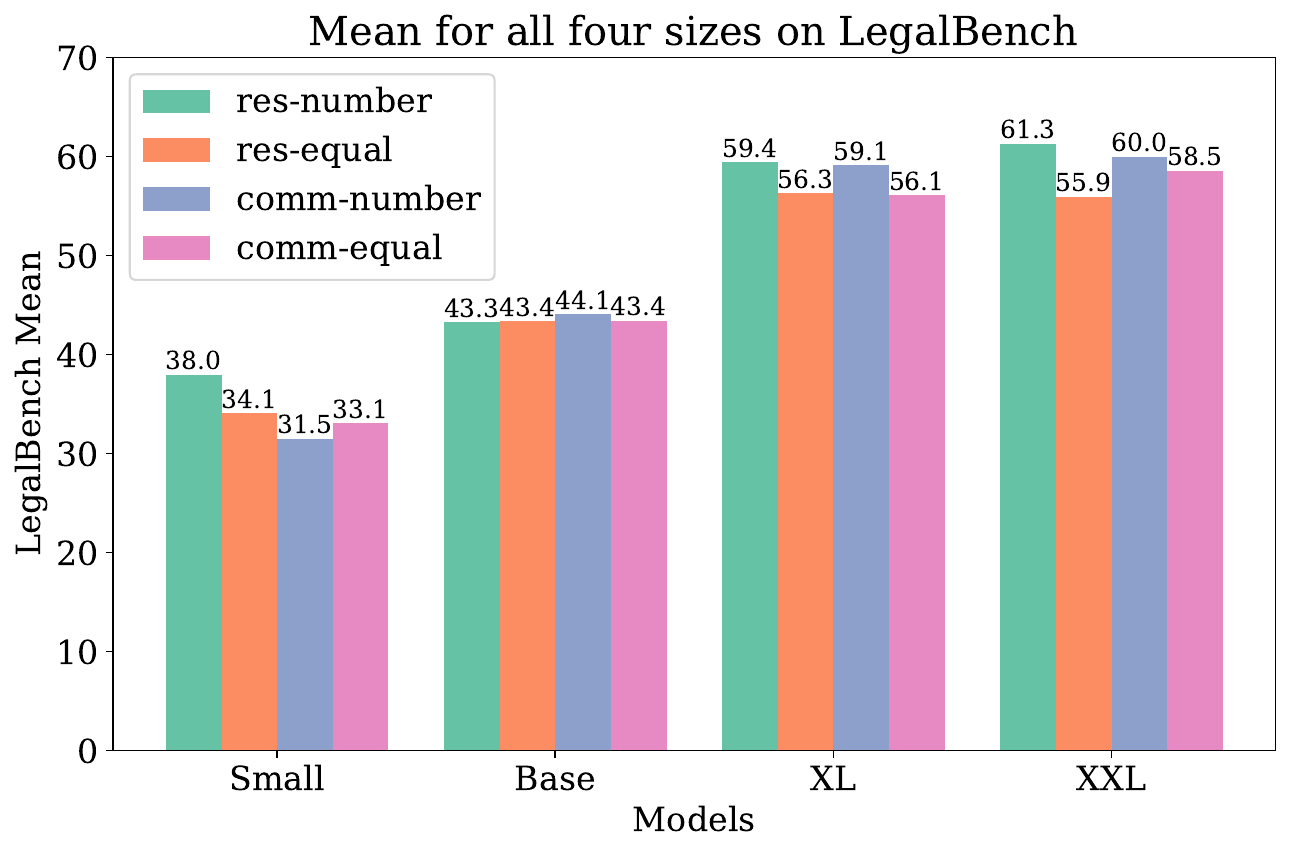}
    \caption{Ablation on sampling style and license on English flan2-lawinstruct from the Flan-T5 checkpoint across sizes. Abbreviations: \emph{res}: licensed for research use (all datasets), \emph{comm}: commercially friendly licensed, \emph{number}: sampling by the number of examples per dataset, \emph{equal}: equally sampling from each dataset}
    \label{fig:sampling_style_license}
    \end{figure}
        
    \emph{Should we sample each dataset equally or rather by the number of examples?}
    $\Rightarrow$ \textbf{Sampling by the number of examples generally leads to better performance.}
    In \Cref{fig:sampling_style_license}, we compare the performance of two sampling styles (equal sampling of each dataset and sampling by the number of examples) across both the research and commercial licensed dataset (detailed results in \Cref{sec:detailed_results} \Cref{tab:licence_sampling_style}). For the XL and XXL sizes, sampling by the number of examples is better than equal weight for datasets for both the research and commercial datasets, although not always statistically significant (XL res $p = 0.049$, XL comm $p = 0.052$, XXL res $p < 0.001$, XXL comm $p = 0.31$). For the Small size, sampling by the number of examples is better for the research dataset ($p < 0.001$) but not for the commercial dataset ($p = 0.099$), while there is no difference for the Base size. 
    By default, we sample by the number of examples in all following experiments unless specified otherwise.

    \subsection{License of Instruction Tuning Datasets}
    \emph{Do we need data licensed non-commercially for good performance?}
    $\Rightarrow$ \textbf{The commercially licensed data seems to be enough for the larger models.}
    In \Cref{fig:sampling_style_license}, we compare the performance of two differently licensed datasets (research and commercial licenses) across both sampling each dataset equally and by the number of examples (detailed results in \Cref{sec:detailed_results} \Cref{tab:licence_sampling_style}). There are fewer datasets available with more permissive licenses allowing for commercial use than for research use (see \Cref{tab:dataset} for details on licenses). Except for Small size ($p < 0.001$), using more diverse data available only for research shows no significant benefit.
    By default, we use the commercially licensed dataset in all subsequent experiments unless specified.

    \subsection{Crosslingual Transfer from Multilingual Data}
    
    \begin{figure}[ht]
    \centering
    \includegraphics[width=0.5\textwidth]{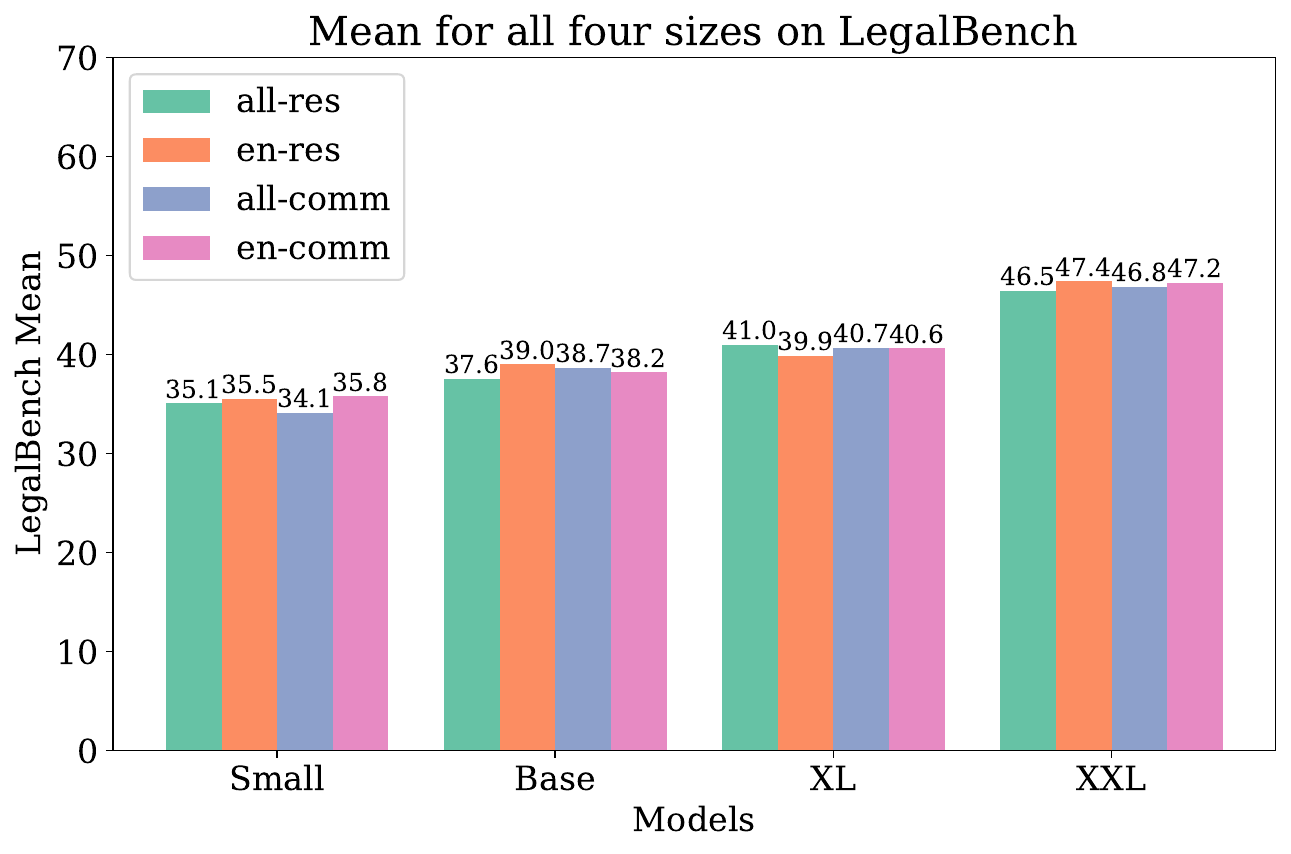}
    \caption{Ablation on the language and license on flan2-lawinstruct from the mT5 checkpoint across all sizes, sampling by the number of examples.}
    \label{fig:language_license}
    \end{figure}
    
    \emph{Is there crosslingual transfer from multilingual data?}
    $\Rightarrow$ \textbf{On the English LegalBench, we do not see any crosslingual transfer.}
    In \Cref{fig:language_license}, we compare the performance of the complete multilingual instruction dataset and the English subset across two differently licensed datasets (research and commercial licenses). We see no statistically significant difference between the multilingual training and the English training. We also see no difference between the differently licensed datasets. This means that just training on the commercial subset is enough. We show detailed results on individual LegalBench categories in \Cref{sec:detailed_results} \Cref{tab:language_license}.
    Per default we use the English dataset in all following experiments unless specified otherwise.

\newpage
\section{Detailed Results}
\label{sec:detailed_results}

\begin{table*}[h]
\centering
\caption{Baseline results on LegalBench.}
\resizebox{\textwidth}{!}{
\begin{tabular}{lrrrrrr}
\toprule
\bf LLM & \bf Issue & \bf Rule & \bf Conclusion & \bf Interpretation & \bf Rhetorical & \bf LegalBench \\
\midrule
Flan-T5 XXL (ours) & 36.1  & 18.8 & 25.2  & 35.1  & 41.1 & 31.3 \\
Flan-T5-XXL \citep{guha_legalbench_2023} & 66.0 & 36.0 & 63.3 & 64.4 & 70.7 & 60.1 \\
LLaMA-2-13B \citep{guha_legalbench_2023} & 50.2 & 37.7 & 59.3 & 50.9 & 54.9 & 50.6 \\
OPT-13B \citep{guha_legalbench_2023} & 52.9 & 28.4 & 45.0 & 45.1 & 43.2 & 42.9 \\
Vicuna-13B-16k \citep{guha_legalbench_2023} & 34.3 & 29.4 & 34.9 & 40.0 & 30.1 & 33.7 \\
WizardLM-13B \citep{guha_legalbench_2023} & 24.1 & 38.0 & 62.6 & 50.9 & 59.8 & 47.1 \\
\midrule
Flan-T5 XL (ours) & 53.5  & 32.1  & 46.8  & 58.7  & 59.6 & 50.1 \\
Flan-T5-XL \citep{guha_legalbench_2023} & 56.8 & 31.7 & 52.1 & 51.4 & 67.4 & 51.9 \\
BLOOM-3B \citep{guha_legalbench_2023} & 47.4 & 20.6 & 45.0 & 45.0 & 36.4 & 38.9 \\
Incite-3B-Instruct \citep{guha_legalbench_2023} & 51.1 & 26.9 & 47.4 & 49.6 & 40.2 & 43.0 \\
OPT-2.7B \citep{guha_legalbench_2023} & 53.7 & 22.2 & 46.0 & 44.4 & 39.8 & 41.2 \\
\midrule
Flan-T5 Base (ours) & 44.7  & 18.0 & 20.9  & 28.9  & 37.0 & 29.9 \\
\midrule
Flan-T5 Small (ours) & 0.3  & 30.4 & 39.8  & 28.2  & 27.7 & 25.3 \\
\bottomrule
\end{tabular}
}
\label{tab:paper_baselines}
\end{table*}

\begin{table*}[h]
\centering
\caption{The T5 and Flan-T5 models finetuned on flan2-lawinstruct in four sizes.}
\resizebox{\textwidth}{!}{
\begin{tabular}{lrrrrrr}
\toprule
\bf LLM & \bf Issue & \bf Rule & \bf Conclusion & \bf Interpretation & \bf Rhetorical & \bf LegalBench \\
\midrule
Small T5 & 45.5 {\small ± 13.2} & 25.0 {\small ± 28.9} & 25.6 {\small ± 27.4} & 18.6 {\small ± 23.6} & 32.9 {\small ± 26.8} & 29.5 {\small ± 10.3} \\
Small Flan-T5 & 25.0 {\small ± 22.0} & 38.1 {\small ± 25.4} & 33.1 {\small ± 24.4} & 20.6 {\small ± 26.4} & 40.7 {\small ± 19.5} & 31.5 {\small ± 8.5} \\
\midrule
Base T5 & 49.8 {\small ± 0.7} & 38.1 {\small ± 25.4} & 34.0 {\small ± 23.3} & 21.3 {\small ± 22.8} & 38.0 {\small ± 19.4} & 36.2 {\small ± 10.2} \\
Base Flan-T5 & 50.3 {\small ± 2.4} & 38.8 {\small ± 25.9} & 34.0 {\small ± 22.4} & 43.0 {\small ± 21.1} & 54.1 {\small ± 13.0} & 44.1 {\small ± 8.2} \\
\midrule
XL T5 & 47.8 {\small ± 12.5} & 37.5 {\small ± 25.0} & 38.2 {\small ± 15.5} & 28.6 {\small ± 25.1} & 49.4 {\small ± 8.1} & 40.3 {\small ± 8.5} \\
XL Flan-T5 & 65.7 {\small ± 15.2} & 45.1 {\small ± 30.3} & 49.0 {\small ± 23.5} & 56.8 {\small ± 18.8} & 79.0 {\small ± 11.4} & 59.1 {\small ± 13.6} \\
\midrule
XXL T5 & 52.7 {\small ± 6.8} & 38.5 {\small ± 25.7} & 50.0 {\small ± 22.8} & 44.9 {\small ± 25.2} & 70.7 {\small ± 20.5} & 51.4 {\small ± 12.1} \\
XXL Flan-T5 & 55.2 {\small ± 23.7} & 46.3 {\small ± 31.6} & 56.1 {\small ± 29.1} & 57.7 {\small ± 19.8} & 84.6 {\small ± 9.6} & 60.0 {\small ± 14.4} \\
\bottomrule
\end{tabular}
}
\label{tab:starting_checkpoint_ift}
\end{table*}

\begin{table*}[h]
\centering
\caption{The Flan-T5 models finetuned on three different data mixtures.}
\resizebox{\textwidth}{!}{
\begin{tabular}{lrrrrrr}
\toprule
\bf LLM & \bf Issue & \bf Rule & \bf Conclusion & \bf Interpretation & \bf Rhetorical & \bf LegalBench \\
\midrule
Small baseline & 0.3 {\small ± 0.7} & 30.4 {\small ± 20.3} & 23.8 {\small ± 25.0} & 16.9 {\small ± 21.1} & 32.8 {\small ± 21.4} & 20.8 {\small ± 13.0} \\
Small lawinstruct & 0.0 {\small ± 0.1} & 15.9 {\small ± 23.9} & 10.7 {\small ± 22.7} & 10.5 {\small ± 19.8} & 18.6 {\small ± 25.7} & 11.1 {\small ± 7.1} \\
Small flan2 & 28.2 {\small ± 22.4} & 37.8 {\small ± 25.3} & 35.1 {\small ± 24.2} & 22.6 {\small ± 23.3} & 40.5 {\small ± 19.4} & 32.8 {\small ± 7.3} \\
Small flan2-lawinstruct & 25.0 {\small ± 22.0} & 38.1 {\small ± 25.4} & 33.1 {\small ± 24.4} & 20.6 {\small ± 26.4} & 40.7 {\small ± 19.5} & 31.5 {\small ± 8.5} \\
\midrule
Base baseline & 44.7 {\small ± 12.4} & 18.0 {\small ± 23.6} & 36.0 {\small ± 23.8} & 15.6 {\small ± 19.9} & 42.7 {\small ± 19.8} & 31.4 {\small ± 13.8} \\
Base lawinstruct & 14.6 {\small ± 14.7} & 22.3 {\small ± 26.3} & 30.2 {\small ± 22.6} & 19.7 {\small ± 26.0} & 17.8 {\small ± 27.4} & 20.9 {\small ± 5.9} \\
Base flan2 & 47.2 {\small ± 4.3} & 37.6 {\small ± 25.0} & 28.6 {\small ± 23.4} & 32.5 {\small ± 21.9} & 54.4 {\small ± 16.3} & 40.0 {\small ± 10.6} \\
Base flan2-lawinstruct & 50.3 {\small ± 2.4} & 38.8 {\small ± 25.9} & 34.0 {\small ± 22.4} & 43.0 {\small ± 21.1} & 54.1 {\small ± 13.0} & 44.1 {\small ± 8.2} \\
\midrule
XL baseline & 53.5 {\small ± 6.0} & 32.1 {\small ± 24.6} & 38.2 {\small ± 22.4} & 49.8 {\small ± 22.6} & 68.1 {\small ± 20.1} & 48.3 {\small ± 14.0} \\
XL lawinstruct & 54.5 {\small ± 7.7} & 30.2 {\small ± 35.1} & 42.9 {\small ± 20.8} & 39.8 {\small ± 30.8} & 63.7 {\small ± 14.1} & 46.2 {\small ± 13.1} \\
XL flan2 & 65.5 {\small ± 14.6} & 40.6 {\small ± 27.7} & 52.0 {\small ± 25.6} & 53.0 {\small ± 21.9} & 74.0 {\small ± 20.8} & 57.0 {\small ± 13.0} \\
XL flan2-lawinstruct & 65.7 {\small ± 15.2} & 45.1 {\small ± 30.3} & 49.0 {\small ± 23.5} & 56.8 {\small ± 18.8} & 79.0 {\small ± 11.4} & 59.1 {\small ± 13.6} \\
\midrule
XXL baseline & 36.1 {\small ± 21.5} & 18.8 {\small ± 24.6} & 39.4 {\small ± 32.1} & 25.7 {\small ± 24.2} & 47.6 {\small ± 14.0} & 33.5 {\small ± 11.4} \\
XXL lawinstruct & 54.1 {\small ± 7.2} & 37.7 {\small ± 27.2} & 53.2 {\small ± 32.6} & 46.7 {\small ± 25.0} & 73.7 {\small ± 15.1} & 53.1 {\small ± 13.3} \\
XXL flan2 & 64.0 {\small ± 12.6} & 44.7 {\small ± 31.4} & 56.4 {\small ± 27.7} & 55.5 {\small ± 20.2} & 81.3 {\small ± 9.7} & 60.4 {\small ± 13.6} \\
XXL flan2-lawinstruct & 55.2 {\small ± 23.7} & 46.3 {\small ± 31.6} & 56.1 {\small ± 29.1} & 57.7 {\small ± 19.8} & 84.6 {\small ± 9.6} & 60.0 {\small ± 14.4} \\
\bottomrule
\end{tabular}
}
\label{tab:data_mixtures}
\end{table*}

\begin{table*}
\centering
\caption{Flan-T5 models finetuned on four different licence-sampling style configurations.}
\resizebox{\textwidth}{!}{
\begin{tabular}{lrrrrrr}
\toprule
\bf LLM & \bf Issue & \bf Rule & \bf Conclusion & \bf Interpretation & \bf Rhetorical & \bf LegalBench \\
\midrule
Small res-number & 50.3 {\small ± 1.3} & 38.2 {\small ± 25.5} & 34.9 {\small ± 25.6} & 21.3 {\small ± 26.6} & 45.3 {\small ± 22.0} & 38.0 {\small ± 11.1} \\
Small res-equal & 34.9 {\small ± 21.2} & 37.5 {\small ± 25.0} & 33.0 {\small ± 25.3} & 21.1 {\small ± 25.1} & 43.8 {\small ± 19.2} & 34.1 {\small ± 8.3} \\
Small comm-number & 25.0 {\small ± 22.0} & 38.1 {\small ± 25.4} & 33.1 {\small ± 24.4} & 20.6 {\small ± 26.4} & 40.7 {\small ± 19.5} & 31.5 {\small ± 8.5} \\
Small comm-equal & 31.6 {\small ± 25.1} & 37.2 {\small ± 24.8} & 33.6 {\small ± 22.8} & 20.2 {\small ± 24.0} & 42.6 {\small ± 21.3} & 33.1 {\small ± 8.3} \\
\midrule
Base res-number & 49.8 {\small ± 3.2} & 38.1 {\small ± 25.4} & 36.0 {\small ± 23.8} & 42.8 {\small ± 21.3} & 49.5 {\small ± 12.1} & 43.3 {\small ± 6.3} \\
Base res-equal & 48.9 {\small ± 3.8} & 39.4 {\small ± 26.3} & 38.4 {\small ± 25.6} & 36.6 {\small ± 19.8} & 53.4 {\small ± 18.3} & 43.4 {\small ± 7.4} \\
Base comm-number & 50.3 {\small ± 2.4} & 38.8 {\small ± 25.9} & 34.0 {\small ± 22.4} & 43.0 {\small ± 21.1} & 54.1 {\small ± 13.0} & 44.1 {\small ± 8.2} \\
Base comm-equal & 49.2 {\small ± 2.9} & 38.5 {\small ± 25.7} & 36.4 {\small ± 20.3} & 40.5 {\small ± 19.8} & 52.6 {\small ± 13.3} & 43.4 {\small ± 7.1} \\
\midrule
XL res-number & 59.9 {\small ± 10.4} & 44.2 {\small ± 29.8} & 53.5 {\small ± 28.0} & 57.1 {\small ± 20.2} & 82.4 {\small ± 11.1} & 59.4 {\small ± 14.2} \\
XL res-equal & 58.2 {\small ± 8.4} & 42.3 {\small ± 28.7} & 46.6 {\small ± 16.8} & 55.4 {\small ± 19.3} & 79.0 {\small ± 11.9} & 56.3 {\small ± 14.3} \\
XL comm-number & 65.7 {\small ± 15.2} & 45.1 {\small ± 30.3} & 49.0 {\small ± 23.5} & 56.8 {\small ± 18.8} & 79.0 {\small ± 11.4} & 59.1 {\small ± 13.6} \\
XL comm-equal & 59.3 {\small ± 10.4} & 40.6 {\small ± 27.2} & 47.7 {\small ± 20.7} & 54.1 {\small ± 20.0} & 78.7 {\small ± 11.9} & 56.1 {\small ± 14.4} \\
\midrule
XXL res-number & 62.9 {\small ± 12.3} & 46.9 {\small ± 31.7} & 57.6 {\small ± 30.2} & 56.7 {\small ± 21.5} & 82.3 {\small ± 9.3} & 61.3 {\small ± 13.1} \\
XXL res-equal & 54.9 {\small ± 6.3} & 43.3 {\small ± 30.1} & 55.5 {\small ± 27.3} & 55.4 {\small ± 19.2} & 70.5 {\small ± 11.6} & 55.9 {\small ± 9.6} \\
XXL comm-number & 55.2 {\small ± 23.7} & 46.3 {\small ± 31.6} & 56.1 {\small ± 29.1} & 57.7 {\small ± 19.8} & 84.6 {\small ± 9.6} & 60.0 {\small ± 14.4} \\
XXL comm-equal & 59.5 {\small ± 13.1} & 45.7 {\small ± 30.0} & 54.8 {\small ± 27.6} & 55.4 {\small ± 19.6} & 77.2 {\small ± 12.3} & 58.5 {\small ± 11.6} \\
\bottomrule
\end{tabular}
}
\label{tab:licence_sampling_style}
\end{table*}

\begin{table*}[h]
\centering
\caption{Flan-T5 models finetuned on four different language-license configurations.}
\resizebox{\textwidth}{!}{
\begin{tabular}{lrrrrrr}
\toprule
\bf LLM & \bf Issue & \bf Rule & \bf Conclusion & \bf Interpretation & \bf Rhetorical & \bf LegalBench \\
\midrule
Small all-res & 46.8 {\small ± 12.2} & 38.2 {\small ± 24.2} & 33.8 {\small ± 22.8} & 20.1 {\small ± 22.0} & 36.5 {\small ± 21.1} & 35.1 {\small ± 9.7} \\
Small en-res & 50.7 {\small ± 5.9} & 37.4 {\small ± 24.9} & 34.0 {\small ± 23.0} & 18.5 {\small ± 22.9} & 37.1 {\small ± 23.0} & 35.5 {\small ± 11.5} \\
Small all-comm & 49.7 {\small ± 2.1} & 38.0 {\small ± 24.1} & 34.0 {\small ± 23.0} & 13.2 {\small ± 19.9} & 35.8 {\small ± 21.8} & 34.1 {\small ± 13.2} \\
Small en-comm & 49.1 {\small ± 13.3} & 37.5 {\small ± 25.0} & 34.4 {\small ± 23.3} & 19.7 {\small ± 23.2} & 38.2 {\small ± 24.2} & 35.8 {\small ± 10.6} \\
\midrule
Base all-res & 51.7 {\small ± 4.4} & 38.7 {\small ± 26.1} & 33.6 {\small ± 22.7} & 22.0 {\small ± 23.6} & 41.8 {\small ± 18.5} & 37.6 {\small ± 10.9} \\
Base en-res & 51.8 {\small ± 5.5} & 37.5 {\small ± 25.0} & 37.1 {\small ± 16.5} & 20.7 {\small ± 22.7} & 48.0 {\small ± 18.1} & 39.0 {\small ± 12.1} \\
Base all-comm & 51.8 {\small ± 5.2} & 38.0 {\small ± 25.4} & 34.3 {\small ± 22.9} & 23.7 {\small ± 24.7} & 45.5 {\small ± 12.6} & 38.7 {\small ± 10.7} \\
Base en-comm & 52.0 {\small ± 3.7} & 37.5 {\small ± 25.0} & 33.2 {\small ± 22.7} & 21.9 {\small ± 21.8} & 46.5 {\small ± 21.2} & 38.2 {\small ± 11.7} \\
\midrule
XL all-res & 49.9 {\small ± 0.9} & 37.5 {\small ± 25.0} & 36.9 {\small ± 18.1} & 28.3 {\small ± 22.9} & 52.2 {\small ± 10.7} & 41.0 {\small ± 9.9} \\
XL en-res & 49.9 {\small ± 0.3} & 37.5 {\small ± 25.0} & 36.6 {\small ± 18.4} & 24.8 {\small ± 25.9} & 50.5 {\small ± 8.6} & 39.9 {\small ± 10.7} \\
XL all-comm & 51.5 {\small ± 2.3} & 37.5 {\small ± 25.0} & 36.9 {\small ± 18.1} & 26.8 {\small ± 24.2} & 50.7 {\small ± 9.4} & 40.7 {\small ± 10.4} \\
XL en-comm & 49.9 {\small ± 1.0} & 37.5 {\small ± 25.0} & 38.3 {\small ± 16.0} & 27.2 {\small ± 24.3} & 50.3 {\small ± 9.8} & 40.6 {\small ± 9.7} \\
\midrule
XXL all-res & 51.5 {\small ± 2.8} & 38.2 {\small ± 24.2} & 40.9 {\small ± 18.5} & 45.3 {\small ± 19.0} & 56.4 {\small ± 10.4} & 46.5 {\small ± 7.5} \\
XXL en-res & 53.4 {\small ± 5.4} & 39.0 {\small ± 24.8} & 40.1 {\small ± 20.5} & 45.4 {\small ± 20.6} & 59.0 {\small ± 9.9} & 47.4 {\small ± 8.7} \\
XXL all-comm & 50.6 {\small ± 1.4} & 38.3 {\small ± 24.3} & 45.2 {\small ± 22.4} & 41.0 {\small ± 20.2} & 58.9 {\small ± 8.7} & 46.8 {\small ± 8.2} \\
XXL en-comm & 52.5 {\small ± 4.1} & 33.3 {\small ± 27.0} & 43.9 {\small ± 24.8} & 47.2 {\small ± 17.8} & 59.2 {\small ± 16.2} & 47.2 {\small ± 9.7} \\
\bottomrule
\end{tabular}
}
\label{tab:language_license}
\end{table*}

\begin{table*}[h]
\centering
\caption{Flan-T5 models finetuned on two different instruction style configurations.}
\resizebox{\textwidth}{!}{
\begin{tabular}{lrrrrrr}
\toprule
\bf LLM & \bf Issue & \bf Rule & \bf Conclusion & \bf Interpretation & \bf Rhetorical & \bf LegalBench \\
\midrule
Small 1-english & 28.3 {\small ± 22.1} & 37.5 {\small ± 25.0} & 35.3 {\small ± 20.2} & 21.8 {\small ± 26.5} & 44.8 {\small ± 17.9} & 33.6 {\small ± 8.8} \\
Small 10-english & 25.0 {\small ± 22.0} & 38.1 {\small ± 25.4} & 33.1 {\small ± 24.4} & 20.6 {\small ± 26.4} & 40.7 {\small ± 19.5} & 31.5 {\small ± 8.5} \\
\midrule
Base 1-english & 51.1 {\small ± 6.2} & 39.0 {\small ± 26.0} & 36.2 {\small ± 21.6} & 43.6 {\small ± 21.2} & 57.6 {\small ± 14.7} & 45.5 {\small ± 8.8} \\
Base 10-english & 50.3 {\small ± 2.4} & 38.8 {\small ± 25.9} & 34.0 {\small ± 22.4} & 43.0 {\small ± 21.1} & 54.1 {\small ± 13.0} & 44.1 {\small ± 8.2} \\
\midrule
XL 1-english & 60.6 {\small ± 11.1} & 42.5 {\small ± 28.8} & 52.1 {\small ± 24.4} & 55.0 {\small ± 18.7} & 81.3 {\small ± 11.1} & 58.3 {\small ± 14.5} \\
XL 10-english & 65.7 {\small ± 15.2} & 45.1 {\small ± 30.3} & 49.0 {\small ± 23.5} & 56.8 {\small ± 18.8} & 79.0 {\small ± 11.4} & 59.1 {\small ± 13.6} \\
\midrule
XXL 1-english & 63.0 {\small ± 13.1} & 43.9 {\small ± 29.7} & 59.0 {\small ± 30.5} & 58.1 {\small ± 20.2} & 80.7 {\small ± 9.9} & 60.9 {\small ± 13.2} \\
XXL 10-english & 55.2 {\small ± 23.7} & 46.3 {\small ± 31.6} & 56.1 {\small ± 29.1} & 57.7 {\small ± 19.8} & 84.6 {\small ± 9.6} & 60.0 {\small ± 14.4} \\
\bottomrule
\end{tabular}
}
\label{tab:instruction_style_english}
\end{table*}

\begin{table*}[h]
\centering
\caption{mT5 models finetuned on three different instruction style configurations.}
\resizebox{\textwidth}{!}{
\begin{tabular}{lrrrrrr}
\toprule
\bf LLM & \bf Issue & \bf Rule & \bf Conclusion & \bf Interpretation & \bf Rhetorical & \bf LegalBench \\
\midrule
Small 1-english & 30.2 {\small ± 20.4} & 39.4 {\small ± 25.1} & 35.0 {\small ± 24.3} & 18.3 {\small ± 24.2} & 37.8 {\small ± 24.4} & 32.2 {\small ± 8.5} \\
Small 10-english & 50.8 {\small ± 3.1} & 38.4 {\small ± 25.7} & 33.8 {\small ± 23.6} & 17.9 {\small ± 23.9} & 36.0 {\small ± 23.0} & 35.4 {\small ± 11.8} \\
Small 10-multi & 46.5 {\small ± 13.4} & 39.4 {\small ± 25.1} & 33.4 {\small ± 23.5} & 18.2 {\small ± 24.2} & 36.9 {\small ± 23.5} & 34.9 {\small ± 10.5} \\
\midrule
Base 1-english & 53.4 {\small ± 5.7} & 37.5 {\small ± 23.8} & 34.7 {\small ± 23.7} & 26.3 {\small ± 23.7} & 44.3 {\small ± 20.0} & 39.2 {\small ± 10.2} \\
Base 10-english & 52.4 {\small ± 5.1} & 37.3 {\small ± 23.6} & 38.0 {\small ± 17.9} & 21.8 {\small ± 23.0} & 41.5 {\small ± 20.5} & 38.2 {\small ± 11.0} \\
Base 10-multi & 51.3 {\small ± 3.2} & 38.0 {\small ± 24.1} & 34.4 {\small ± 22.7} & 29.6 {\small ± 21.2} & 41.7 {\small ± 18.1} & 39.0 {\small ± 8.2} \\
\midrule
XL 1-english & 51.7 {\small ± 3.4} & 38.0 {\small ± 24.1} & 36.9 {\small ± 18.1} & 36.3 {\small ± 21.7} & 50.9 {\small ± 8.9} & 42.7 {\small ± 7.8} \\
XL 10-english & 43.6 {\small ± 16.5} & 38.0 {\small ± 24.1} & 36.9 {\small ± 18.1} & 30.9 {\small ± 20.0} & 45.6 {\small ± 13.8} & 39.0 {\small ± 5.8} \\
XL 10-multi & 51.2 {\small ± 3.3} & 38.0 {\small ± 24.1} & 36.9 {\small ± 18.1} & 31.1 {\small ± 25.4} & 54.8 {\small ± 12.9} & 42.4 {\small ± 10.1} \\
\bottomrule
\end{tabular}
}
\label{tab:instruction_style_multi}
\end{table*}

\begin{table*}[h]
\centering
\caption{Flan-T5 Small models with different domain adaptation strategies (amount of IFT data during continued pretraining). 1-IFT-to-X-PRE means that for every X pretraining examples we mix in one instruction example. ONLY-PRE means we did not mix in any instruction examples.}
\resizebox{\textwidth}{!}{
\begin{tabular}{lrrrrrr}
\toprule
\bf LLM & \bf Issue & \bf Rule & \bf Conclusion & \bf Interpretation & \bf Rhetorical & \bf LegalBench \\
\midrule
Baseline & 0.3 {\small ± 0.7} & 30.4 {\small ± 20.3} & 39.8 {\small ± 20.8} & 28.2 {\small ± 21.6} & 27.7 {\small ± 21.9} & 25.3 {\small ± 13.2} \\
IFT & 25.0 {\small ± 22.0} & 38.1 {\small ± 25.4} & 43.0 {\small ± 17.1} & 36.1 {\small ± 26.5} & 32.6 {\small ± 24.2} & 34.9 {\small ± 6.0} \\
\midrule
1-IFT-to-200-PRE+IFT 10K & 50.6 {\small ± 4.2} & 38.2 {\small ± 25.6} & 44.3 {\small ± 15.6} & 33.8 {\small ± 23.3} & 33.7 {\small ± 23.8} & 40.1 {\small ± 6.5} \\
1-IFT-to-200-PRE+IFT 20K & 50.8 {\small ± 2.2} & 37.9 {\small ± 25.3} & 44.4 {\small ± 15.7} & 35.5 {\small ± 25.1} & 31.9 {\small ± 24.0} & 40.1 {\small ± 6.7} \\
1-IFT-to-200-PRE+IFT 30K & 42.2 {\small ± 16.2} & 37.3 {\small ± 24.9} & 39.8 {\small ± 19.4} & 34.3 {\small ± 23.7} & 32.4 {\small ± 23.5} & 37.2 {\small ± 3.6} \\
1-IFT-to-200-PRE+IFT 40K & 45.8 {\small ± 10.8} & 37.7 {\small ± 25.2} & 39.7 {\small ± 20.8} & 35.1 {\small ± 24.4} & 33.4 {\small ± 24.0} & 38.3 {\small ± 4.3} \\
1-IFT-to-200-PRE+IFT 50K & 47.0 {\small ± 8.8} & 37.4 {\small ± 24.9} & 38.9 {\small ± 20.7} & 35.6 {\small ± 24.6} & 34.1 {\small ± 21.0} & 38.6 {\small ± 4.5} \\
1-IFT-to-200-PRE+IFT 60K & 50.0 {\small ± 0.4} & 37.1 {\small ± 24.7} & 39.3 {\small ± 18.7} & 34.7 {\small ± 23.3} & 33.8 {\small ± 21.7} & 39.0 {\small ± 5.8} \\
1-IFT-to-200-PRE+IFT 70K & 41.4 {\small ± 16.9} & 38.4 {\small ± 25.6} & 38.8 {\small ± 21.1} & 34.0 {\small ± 22.7} & 33.8 {\small ± 22.9} & 37.3 {\small ± 2.9} \\
1-IFT-to-200-PRE+IFT 80K & 51.8 {\small ± 3.8} & 38.2 {\small ± 25.5} & 38.5 {\small ± 20.9} & 36.2 {\small ± 22.6} & 33.4 {\small ± 21.5} & 39.6 {\small ± 6.4} \\
1-IFT-to-200-PRE+IFT 90K & 42.4 {\small ± 16.7} & 37.9 {\small ± 25.3} & 39.7 {\small ± 20.3} & 35.8 {\small ± 23.5} & 34.1 {\small ± 22.2} & 38.0 {\small ± 2.9} \\
\midrule
1-IFT-to-1000-PRE+IFT 10K & 42.3 {\small ± 16.1} & 38.1 {\small ± 25.4} & 43.9 {\small ± 15.0} & 33.6 {\small ± 23.8} & 32.7 {\small ± 24.5} & 38.1 {\small ± 4.5} \\
1-IFT-to-1000-PRE+IFT 20K & 41.7 {\small ± 20.5} & 37.0 {\small ± 24.7} & 42.9 {\small ± 16.6} & 33.1 {\small ± 23.4} & 33.0 {\small ± 24.6} & 37.5 {\small ± 4.2} \\
1-IFT-to-1000-PRE+IFT 30K & 49.9 {\small ± 0.4} & 37.8 {\small ± 25.3} & 40.3 {\small ± 17.7} & 34.3 {\small ± 24.2} & 32.4 {\small ± 23.5} & 38.9 {\small ± 6.1} \\
1-IFT-to-1000-PRE+IFT 40K & 51.4 {\small ± 2.7} & 37.8 {\small ± 25.2} & 38.9 {\small ± 20.6} & 34.7 {\small ± 24.4} & 33.0 {\small ± 22.5} & 39.2 {\small ± 6.5} \\
1-IFT-to-1000-PRE+IFT 50K & 51.6 {\small ± 2.7} & 37.7 {\small ± 25.2} & 39.8 {\small ± 18.4} & 33.7 {\small ± 23.3} & 33.8 {\small ± 22.4} & 39.3 {\small ± 6.6} \\
1-IFT-to-1000-PRE+IFT 60K & 50.0 {\small ± 0.6} & 37.5 {\small ± 25.0} & 40.5 {\small ± 20.2} & 34.4 {\small ± 23.5} & 33.2 {\small ± 22.4} & 39.1 {\small ± 6.0} \\
1-IFT-to-1000-PRE+IFT 70K & 50.3 {\small ± 1.4} & 37.3 {\small ± 24.9} & 43.1 {\small ± 17.1} & 34.6 {\small ± 24.6} & 33.1 {\small ± 22.4} & 39.7 {\small ± 6.3} \\
1-IFT-to-1000-PRE+IFT 80K & 50.6 {\small ± 1.5} & 37.7 {\small ± 25.2} & 43.0 {\small ± 17.4} & 34.0 {\small ± 23.1} & 32.9 {\small ± 23.0} & 39.6 {\small ± 6.5} \\
1-IFT-to-1000-PRE+IFT 90K & 51.6 {\small ± 2.6} & 37.0 {\small ± 24.7} & 40.2 {\small ± 19.2} & 34.4 {\small ± 24.8} & 32.9 {\small ± 21.4} & 39.2 {\small ± 6.7} \\
\midrule
1-IFT-to-10000-PRE+IFT 10K & 46.0 {\small ± 12.1} & 38.0 {\small ± 25.4} & 44.4 {\small ± 15.5} & 33.5 {\small ± 23.3} & 33.8 {\small ± 24.3} & 39.1 {\small ± 5.2} \\
1-IFT-to-10000-PRE+IFT 20K & 50.5 {\small ± 1.4} & 37.9 {\small ± 25.3} & 44.3 {\small ± 15.4} & 34.9 {\small ± 25.2} & 32.1 {\small ± 24.0} & 39.9 {\small ± 6.7} \\
1-IFT-to-10000-PRE+IFT 30K & 51.3 {\small ± 4.0} & 38.2 {\small ± 25.5} & 40.5 {\small ± 18.1} & 33.6 {\small ± 23.3} & 34.7 {\small ± 26.5} & 39.7 {\small ± 6.3} \\
1-IFT-to-10000-PRE+IFT 40K & 52.3 {\small ± 4.4} & 38.9 {\small ± 26.1} & 38.8 {\small ± 19.8} & 33.2 {\small ± 23.0} & 33.6 {\small ± 25.3} & 39.4 {\small ± 6.9} \\
1-IFT-to-10000-PRE+IFT 50K & 47.3 {\small ± 12.3} & 37.6 {\small ± 25.1} & 41.5 {\small ± 17.2} & 35.1 {\small ± 24.4} & 32.8 {\small ± 22.2} & 38.8 {\small ± 5.1} \\
1-IFT-to-10000-PRE+IFT 60K & 49.4 {\small ± 2.7} & 38.1 {\small ± 25.5} & 39.0 {\small ± 20.6} & 35.3 {\small ± 24.3} & 32.2 {\small ± 23.2} & 38.8 {\small ± 5.8} \\
1-IFT-to-10000-PRE+IFT 70K & 49.2 {\small ± 13.9} & 37.7 {\small ± 25.2} & 42.1 {\small ± 16.2} & 33.2 {\small ± 23.1} & 33.8 {\small ± 24.3} & 39.2 {\small ± 5.9} \\
1-IFT-to-10000-PRE+IFT 80K & 51.4 {\small ± 7.0} & 37.5 {\small ± 25.0} & 42.5 {\small ± 16.0} & 33.5 {\small ± 22.4} & 32.7 {\small ± 22.4} & 39.5 {\small ± 6.9} \\
1-IFT-to-10000-PRE+IFT 90K & 44.1 {\small ± 20.2} & 37.5 {\small ± 25.0} & 43.0 {\small ± 16.4} & 33.6 {\small ± 22.3} & 33.0 {\small ± 21.9} & 38.2 {\small ± 4.6} \\
\midrule
ONLY-PRE+IFT 10K & 51.1 {\small ± 3.1} & 37.9 {\small ± 25.3} & 44.9 {\small ± 16.9} & 33.8 {\small ± 23.6} & 34.6 {\small ± 24.7} & 40.5 {\small ± 6.6} \\
ONLY-PRE+IFT 20K & 51.4 {\small ± 4.4} & 38.1 {\small ± 25.5} & 43.9 {\small ± 14.0} & 34.1 {\small ± 25.1} & 33.2 {\small ± 25.3} & 40.2 {\small ± 6.8} \\
ONLY-PRE+IFT 30K & 43.0 {\small ± 17.8} & 37.9 {\small ± 25.4} & 42.2 {\small ± 16.2} & 35.1 {\small ± 25.6} & 32.4 {\small ± 23.6} & 38.1 {\small ± 4.1} \\
ONLY-PRE+IFT 40K & 47.1 {\small ± 12.5} & 38.4 {\small ± 25.6} & 42.5 {\small ± 16.6} & 34.9 {\small ± 25.0} & 32.9 {\small ± 24.5} & 39.2 {\small ± 5.1} \\
ONLY-PRE+IFT 50K & 42.0 {\small ± 19.2} & 37.8 {\small ± 25.2} & 42.3 {\small ± 17.4} & 34.8 {\small ± 25.1} & 32.4 {\small ± 23.3} & 37.8 {\small ± 3.9} \\
ONLY-PRE+IFT 60K & 50.6 {\small ± 2.1} & 37.9 {\small ± 25.3} & 43.0 {\small ± 16.0} & 35.6 {\small ± 25.0} & 32.6 {\small ± 22.9} & 39.9 {\small ± 6.3} \\
ONLY-PRE+IFT 70K & 48.6 {\small ± 7.0} & 38.1 {\small ± 25.4} & 42.6 {\small ± 17.0} & 34.8 {\small ± 24.3} & 32.6 {\small ± 24.0} & 39.4 {\small ± 5.7} \\
ONLY-PRE+IFT 80K & 51.2 {\small ± 3.4} & 37.5 {\small ± 25.0} & 43.7 {\small ± 17.2} & 33.2 {\small ± 23.1} & 34.0 {\small ± 25.7} & 39.9 {\small ± 6.7} \\
ONLY-PRE+IFT 90K & 51.5 {\small ± 3.7} & 37.5 {\small ± 25.0} & 40.7 {\small ± 17.5} & 34.7 {\small ± 21.8} & 33.7 {\small ± 24.4} & 39.6 {\small ± 6.4} \\
\bottomrule
\end{tabular}
}
\label{tab:domain_adaptation_small}
\end{table*}

\begin{table*}[h]
\centering
\caption{Flan-T5 Base models with different domain adaptation strategies (amount of IFT data during continued pretraining). 1-IFT-to-X-PRE means that for every X pretraining examples we mix in one instruction example. ONLY-PRE means we did not mix in any instruction examples.}
\resizebox{\textwidth}{!}{
\begin{tabular}{lrrrrrr}
\toprule
\bf LLM & \bf Issue & \bf Rule & \bf Conclusion & \bf Interpretation & \bf Rhetorical & \bf LegalBench \\
\midrule
Baseline & 44.7 {\small ± 12.4} & 18.0 {\small ± 23.6} & 20.9 {\small ± 24.8} & 28.9 {\small ± 21.2} & 37.0 {\small ± 21.3} & 29.9 {\small ± 9.9} \\
IFT & 50.3 {\small ± 2.4} & 38.8 {\small ± 25.9} & 40.5 {\small ± 15.7} & 49.5 {\small ± 19.1} & 45.2 {\small ± 22.0} & 44.9 {\small ± 4.6} \\
\midrule
1-IFT-to-200-PRE+IFT 10K & 50.5 {\small ± 3.2} & 37.3 {\small ± 24.9} & 40.7 {\small ± 16.6} & 47.7 {\small ± 17.7} & 49.7 {\small ± 20.8} & 45.2 {\small ± 5.2} \\
1-IFT-to-200-PRE+IFT 20K & 50.4 {\small ± 2.2} & 37.8 {\small ± 25.2} & 40.9 {\small ± 14.2} & 48.4 {\small ± 15.9} & 46.2 {\small ± 24.7} & 44.7 {\small ± 4.7} \\
1-IFT-to-200-PRE+IFT 30K & 49.9 {\small ± 2.6} & 37.7 {\small ± 25.2} & 41.2 {\small ± 14.1} & 45.3 {\small ± 16.1} & 48.4 {\small ± 20.0} & 44.5 {\small ± 4.5} \\
1-IFT-to-200-PRE+IFT 40K & 49.4 {\small ± 4.3} & 37.8 {\small ± 25.2} & 40.4 {\small ± 15.5} & 47.8 {\small ± 17.2} & 49.0 {\small ± 20.9} & 44.9 {\small ± 4.8} \\
1-IFT-to-200-PRE+IFT 50K & 51.2 {\small ± 3.9} & 37.7 {\small ± 25.2} & 41.2 {\small ± 12.7} & 45.0 {\small ± 16.0} & 49.1 {\small ± 20.1} & 44.8 {\small ± 4.9} \\
1-IFT-to-200-PRE+IFT 60K & 50.1 {\small ± 0.9} & 37.6 {\small ± 25.1} & 45.1 {\small ± 13.0} & 44.2 {\small ± 16.0} & 45.2 {\small ± 18.9} & 44.4 {\small ± 4.0} \\
1-IFT-to-200-PRE+IFT 70K & 51.1 {\small ± 2.7} & 37.6 {\small ± 25.0} & 43.4 {\small ± 13.6} & 45.1 {\small ± 15.4} & 46.5 {\small ± 21.0} & 44.7 {\small ± 4.4} \\
1-IFT-to-200-PRE+IFT 80K & 50.4 {\small ± 2.3} & 37.7 {\small ± 25.2} & 42.2 {\small ± 15.9} & 45.2 {\small ± 15.7} & 44.8 {\small ± 22.7} & 44.1 {\small ± 4.1} \\
1-IFT-to-200-PRE+IFT 90K & 51.4 {\small ± 3.6} & 37.7 {\small ± 25.2} & 41.6 {\small ± 14.5} & 42.9 {\small ± 19.0} & 43.2 {\small ± 21.6} & 43.4 {\small ± 4.5} \\
\midrule
1-IFT-to-1000-PRE+IFT 10K & 46.8 {\small ± 4.8} & 38.5 {\small ± 25.7} & 43.9 {\small ± 13.7} & 47.6 {\small ± 16.6} & 45.9 {\small ± 18.0} & 44.5 {\small ± 3.2} \\
1-IFT-to-1000-PRE+IFT 20K & 50.1 {\small ± 2.0} & 37.8 {\small ± 25.2} & 43.2 {\small ± 15.0} & 46.7 {\small ± 15.9} & 48.2 {\small ± 24.9} & 45.2 {\small ± 4.3} \\
1-IFT-to-1000-PRE+IFT 30K & 50.8 {\small ± 3.3} & 38.9 {\small ± 26.0} & 42.3 {\small ± 15.9} & 49.9 {\small ± 17.6} & 50.4 {\small ± 21.4} & 46.5 {\small ± 4.9} \\
1-IFT-to-1000-PRE+IFT 40K & 50.1 {\small ± 0.7} & 38.4 {\small ± 25.7} & 45.1 {\small ± 12.0} & 46.6 {\small ± 16.2} & 48.0 {\small ± 21.4} & 45.7 {\small ± 4.0} \\
1-IFT-to-1000-PRE+IFT 50K & 51.1 {\small ± 3.0} & 37.7 {\small ± 25.1} & 41.9 {\small ± 13.8} & 48.0 {\small ± 19.3} & 50.1 {\small ± 20.5} & 45.8 {\small ± 5.1} \\
1-IFT-to-1000-PRE+IFT 60K & 49.9 {\small ± 2.3} & 37.7 {\small ± 25.1} & 44.2 {\small ± 15.7} & 46.1 {\small ± 18.3} & 49.7 {\small ± 22.1} & 45.5 {\small ± 4.5} \\
1-IFT-to-1000-PRE+IFT 70K & 50.5 {\small ± 1.5} & 38.5 {\small ± 25.7} & 44.9 {\small ± 16.8} & 47.9 {\small ± 15.9} & 49.8 {\small ± 19.2} & 46.3 {\small ± 4.4} \\
1-IFT-to-1000-PRE+IFT 80K & 50.6 {\small ± 2.5} & 37.9 {\small ± 25.2} & 42.4 {\small ± 16.6} & 48.8 {\small ± 19.2} & 48.7 {\small ± 22.8} & 45.7 {\small ± 4.8} \\
1-IFT-to-1000-PRE+IFT 90K & 50.8 {\small ± 4.2} & 37.8 {\small ± 25.2} & 43.4 {\small ± 15.7} & 45.9 {\small ± 16.9} & 47.8 {\small ± 22.0} & 45.1 {\small ± 4.4} \\
\midrule
1-IFT-to-10000-PRE+IFT 10K & 48.8 {\small ± 4.1} & 38.1 {\small ± 25.4} & 43.6 {\small ± 13.4} & 47.4 {\small ± 16.4} & 47.7 {\small ± 19.6} & 45.1 {\small ± 3.9} \\
1-IFT-to-10000-PRE+IFT 20K & 50.0 {\small ± 2.9} & 37.7 {\small ± 25.1} & 41.5 {\small ± 13.6} & 47.2 {\small ± 18.4} & 52.0 {\small ± 20.8} & 45.7 {\small ± 5.3} \\
1-IFT-to-10000-PRE+IFT 30K & 50.5 {\small ± 4.6} & 38.4 {\small ± 25.6} & 44.3 {\small ± 14.6} & 48.4 {\small ± 17.3} & 51.5 {\small ± 20.7} & 46.6 {\small ± 4.8} \\
1-IFT-to-10000-PRE+IFT 40K & 50.2 {\small ± 2.9} & 37.7 {\small ± 25.1} & 42.4 {\small ± 16.4} & 45.6 {\small ± 16.8} & 49.2 {\small ± 20.7} & 45.0 {\small ± 4.6} \\
1-IFT-to-10000-PRE+IFT 50K & 50.3 {\small ± 2.0} & 37.4 {\small ± 24.9} & 41.8 {\small ± 16.2} & 45.8 {\small ± 17.7} & 49.3 {\small ± 21.7} & 44.9 {\small ± 4.8} \\
1-IFT-to-10000-PRE+IFT 60K & 49.6 {\small ± 4.5} & 37.6 {\small ± 25.1} & 43.7 {\small ± 17.3} & 43.1 {\small ± 19.3} & 48.4 {\small ± 22.0} & 44.5 {\small ± 4.3} \\
1-IFT-to-10000-PRE+IFT 70K & 49.6 {\small ± 2.9} & 37.7 {\small ± 25.1} & 46.4 {\small ± 16.0} & 46.9 {\small ± 18.7} & 50.5 {\small ± 22.2} & 46.2 {\small ± 4.5} \\
1-IFT-to-10000-PRE+IFT 80K & 49.7 {\small ± 3.0} & 37.7 {\small ± 25.2} & 45.1 {\small ± 12.2} & 41.1 {\small ± 18.4} & 47.7 {\small ± 23.7} & 44.2 {\small ± 4.3} \\
1-IFT-to-10000-PRE+IFT 90K & 50.0 {\small ± 1.8} & 37.2 {\small ± 24.8} & 40.6 {\small ± 14.5} & 41.8 {\small ± 20.0} & 45.3 {\small ± 22.3} & 43.0 {\small ± 4.4} \\
\midrule
ONLY-PRE+IFT 10K & 50.7 {\small ± 2.7} & 37.2 {\small ± 24.8} & 42.0 {\small ± 16.3} & 48.0 {\small ± 18.6} & 47.6 {\small ± 20.8} & 45.1 {\small ± 4.9} \\
ONLY-PRE+IFT 20K & 50.1 {\small ± 2.6} & 38.2 {\small ± 25.5} & 41.1 {\small ± 13.7} & 45.0 {\small ± 19.7} & 46.7 {\small ± 25.7} & 44.2 {\small ± 4.2} \\
ONLY-PRE+IFT 30K & 50.7 {\small ± 3.6} & 38.0 {\small ± 25.3} & 43.3 {\small ± 15.3} & 44.6 {\small ± 19.0} & 48.3 {\small ± 21.6} & 45.0 {\small ± 4.4} \\
ONLY-PRE+IFT 40K & 50.4 {\small ± 3.8} & 38.4 {\small ± 25.6} & 41.9 {\small ± 14.5} & 47.4 {\small ± 17.4} & 46.8 {\small ± 21.4} & 45.0 {\small ± 4.3} \\
ONLY-PRE+IFT 50K & 50.6 {\small ± 2.5} & 37.5 {\small ± 25.0} & 41.1 {\small ± 12.8} & 44.5 {\small ± 18.6} & 48.2 {\small ± 21.6} & 44.4 {\small ± 4.7} \\
ONLY-PRE+IFT 60K & 49.6 {\small ± 3.4} & 37.6 {\small ± 25.1} & 40.4 {\small ± 15.5} & 47.2 {\small ± 16.7} & 46.3 {\small ± 21.0} & 44.2 {\small ± 4.5} \\
ONLY-PRE+IFT 70K & 50.6 {\small ± 1.9} & 38.4 {\small ± 25.6} & 41.7 {\small ± 13.2} & 46.1 {\small ± 18.7} & 45.5 {\small ± 21.9} & 44.4 {\small ± 4.2} \\
ONLY-PRE+IFT 80K & 51.0 {\small ± 3.1} & 39.2 {\small ± 26.3} & 42.2 {\small ± 15.7} & 46.8 {\small ± 18.0} & 45.3 {\small ± 21.9} & 44.9 {\small ± 4.0} \\
ONLY-PRE+IFT 90K & 50.5 {\small ± 3.8} & 37.4 {\small ± 25.0} & 44.3 {\small ± 14.7} & 43.2 {\small ± 18.1} & 44.4 {\small ± 22.5} & 44.0 {\small ± 4.1} \\
\bottomrule
\end{tabular}
}
\label{tab:domain_adaptation_base}
\end{table*}

\begin{table*}[h]
\centering
\caption{Flan-T5 XL models with different domain adaptation strategies (amount of IFT data during continued pretraining). 1-IFT-to-X-PRE means that for every X pretraining examples we mix in one instruction example. ONLY-PRE means we did not mix in any instruction examples.}
\resizebox{\textwidth}{!}{
\begin{tabular}{lrrrrrr}
\toprule
\bf LLM & \bf Issue & \bf Rule & \bf Conclusion & \bf Interpretation & \bf Rhetorical & \bf LegalBench \\
\midrule
Baseline & 53.5 {\small ± 6.0} & 32.1 {\small ± 24.6} & 46.8 {\small ± 15.6} & 58.7 {\small ± 21.3} & 59.6 {\small ± 25.6} & 50.1 {\small ± 10.1} \\
IFT & 65.7 {\small ± 15.2} & 45.1 {\small ± 30.3} & 49.5 {\small ± 14.2} & 61.7 {\small ± 17.1} & 68.6 {\small ± 24.1} & 58.1 {\small ± 9.2} \\
\midrule
1-IFT-to-200-PRE+IFT 10K & 56.7 {\small ± 6.9} & 41.8 {\small ± 28.1} & 55.2 {\small ± 16.9} & 62.1 {\small ± 18.6} & 66.8 {\small ± 23.7} & 56.5 {\small ± 8.4} \\
1-IFT-to-200-PRE+IFT 20K & 63.4 {\small ± 13.8} & 44.2 {\small ± 29.8} & 52.5 {\small ± 17.4} & 58.7 {\small ± 16.8} & 67.0 {\small ± 23.2} & 57.2 {\small ± 8.1} \\
1-IFT-to-200-PRE+IFT 30K & 58.7 {\small ± 10.3} & 43.6 {\small ± 29.3} & 56.3 {\small ± 18.2} & 60.2 {\small ± 18.4} & 67.9 {\small ± 24.5} & 57.3 {\small ± 7.9} \\
1-IFT-to-200-PRE+IFT 40K & 58.4 {\small ± 9.7} & 42.3 {\small ± 28.2} & 54.3 {\small ± 15.2} & 61.2 {\small ± 18.8} & 67.5 {\small ± 23.6} & 56.7 {\small ± 8.4} \\
1-IFT-to-200-PRE+IFT 50K & 61.4 {\small ± 13.3} & 42.2 {\small ± 28.3} & 51.8 {\small ± 16.3} & 59.4 {\small ± 17.9} & 67.3 {\small ± 23.6} & 56.4 {\small ± 8.7} \\
1-IFT-to-200-PRE+IFT 60K & 57.5 {\small ± 8.7} & 43.6 {\small ± 29.2} & 53.5 {\small ± 15.8} & 60.3 {\small ± 17.5} & 68.2 {\small ± 23.5} & 56.6 {\small ± 8.1} \\
1-IFT-to-200-PRE+IFT 70K & 58.3 {\small ± 10.2} & 43.1 {\small ± 28.8} & 54.3 {\small ± 17.9} & 58.8 {\small ± 18.2} & 67.6 {\small ± 22.6} & 56.4 {\small ± 8.0} \\
1-IFT-to-200-PRE+IFT 80K & 58.9 {\small ± 11.0} & 44.9 {\small ± 30.0} & 51.1 {\small ± 13.2} & 59.8 {\small ± 17.3} & 68.5 {\small ± 23.3} & 56.6 {\small ± 8.1} \\
1-IFT-to-200-PRE+IFT 90K & 55.2 {\small ± 6.9} & 44.4 {\small ± 30.1} & 51.7 {\small ± 15.6} & 57.9 {\small ± 17.0} & 67.7 {\small ± 24.3} & 55.4 {\small ± 7.6} \\
\midrule
1-IFT-to-1000-PRE+IFT 10K & 61.3 {\small ± 11.8} & 41.8 {\small ± 28.0} & 53.4 {\small ± 16.1} & 60.9 {\small ± 18.8} & 67.0 {\small ± 23.1} & 56.9 {\small ± 8.7} \\
1-IFT-to-1000-PRE+IFT 20K & 63.3 {\small ± 13.7} & 44.3 {\small ± 29.6} & 52.2 {\small ± 17.4} & 60.7 {\small ± 17.5} & 67.3 {\small ± 24.6} & 57.6 {\small ± 8.3} \\
1-IFT-to-1000-PRE+IFT 30K & 58.3 {\small ± 9.8} & 43.4 {\small ± 29.2} & 54.4 {\small ± 17.1} & 61.3 {\small ± 20.4} & 70.2 {\small ± 25.4} & 57.5 {\small ± 8.8} \\
1-IFT-to-1000-PRE+IFT 40K & 62.5 {\small ± 13.2} & 45.6 {\small ± 30.6} & 51.3 {\small ± 17.5} & 60.1 {\small ± 18.9} & 68.0 {\small ± 25.6} & 57.5 {\small ± 8.0} \\
1-IFT-to-1000-PRE+IFT 50K & 56.8 {\small ± 7.5} & 44.7 {\small ± 30.2} & 51.5 {\small ± 14.5} & 58.9 {\small ± 16.9} & 69.7 {\small ± 24.9} & 56.3 {\small ± 8.3} \\
1-IFT-to-1000-PRE+IFT 60K & 54.4 {\small ± 5.3} & 42.2 {\small ± 28.2} & 52.7 {\small ± 16.3} & 59.9 {\small ± 17.8} & 67.1 {\small ± 23.5} & 55.2 {\small ± 8.2} \\
1-IFT-to-1000-PRE+IFT 70K & 59.7 {\small ± 10.8} & 44.1 {\small ± 29.5} & 54.5 {\small ± 17.3} & 59.4 {\small ± 17.6} & 67.4 {\small ± 23.4} & 57.0 {\small ± 7.7} \\
1-IFT-to-1000-PRE+IFT 80K & 59.8 {\small ± 11.2} & 41.6 {\small ± 27.9} & 52.8 {\small ± 17.2} & 63.5 {\small ± 19.8} & 67.3 {\small ± 24.5} & 57.0 {\small ± 9.0} \\
1-IFT-to-1000-PRE+IFT 90K & 60.3 {\small ± 10.6} & 44.3 {\small ± 29.7} & 50.5 {\small ± 15.4} & 57.3 {\small ± 15.9} & 67.3 {\small ± 23.1} & 55.9 {\small ± 8.0} \\
\midrule
1-IFT-to-10000-PRE+IFT 10K & 60.0 {\small ± 10.2} & 42.3 {\small ± 28.4} & 52.7 {\small ± 16.0} & 61.6 {\small ± 18.3} & 68.0 {\small ± 22.8} & 56.9 {\small ± 8.8} \\
1-IFT-to-10000-PRE+IFT 20K & 59.5 {\small ± 11.0} & 42.6 {\small ± 28.5} & 52.5 {\small ± 15.7} & 61.6 {\small ± 18.0} & 68.1 {\small ± 25.0} & 56.9 {\small ± 8.7} \\
1-IFT-to-10000-PRE+IFT 30K & 62.2 {\small ± 12.2} & 42.3 {\small ± 28.5} & 53.6 {\small ± 16.7} & 62.5 {\small ± 20.1} & 69.2 {\small ± 25.2} & 57.9 {\small ± 9.3} \\
1-IFT-to-10000-PRE+IFT 40K & 59.7 {\small ± 10.1} & 43.6 {\small ± 29.2} & 53.1 {\small ± 15.9} & 62.6 {\small ± 18.9} & 67.6 {\small ± 23.1} & 57.3 {\small ± 8.3} \\
1-IFT-to-10000-PRE+IFT 50K & 58.8 {\small ± 8.9} & 42.9 {\small ± 29.1} & 52.5 {\small ± 16.9} & 61.1 {\small ± 17.9} & 64.6 {\small ± 25.0} & 56.0 {\small ± 7.6} \\
1-IFT-to-10000-PRE+IFT 60K & 55.3 {\small ± 5.6} & 42.1 {\small ± 28.3} & 52.1 {\small ± 16.6} & 59.1 {\small ± 19.0} & 66.4 {\small ± 23.1} & 55.0 {\small ± 8.0} \\
1-IFT-to-10000-PRE+IFT 70K & 60.3 {\small ± 10.0} & 43.6 {\small ± 29.5} & 51.8 {\small ± 16.8} & 61.2 {\small ± 18.5} & 69.0 {\small ± 24.7} & 57.2 {\small ± 8.7} \\
1-IFT-to-10000-PRE+IFT 80K & 64.7 {\small ± 13.9} & 44.4 {\small ± 29.9} & 50.8 {\small ± 16.9} & 58.4 {\small ± 17.1} & 70.4 {\small ± 25.8} & 57.8 {\small ± 9.3} \\
1-IFT-to-10000-PRE+IFT 90K & 63.3 {\small ± 13.3} & 44.8 {\small ± 30.2} & 51.9 {\small ± 16.3} & 58.7 {\small ± 16.6} & 68.2 {\small ± 25.1} & 57.4 {\small ± 8.3} \\
\midrule
ONLY-PRE+IFT 10K & 62.8 {\small ± 13.6} & 44.3 {\small ± 29.8} & 52.0 {\small ± 16.7} & 58.9 {\small ± 16.2} & 68.2 {\small ± 23.9} & 57.2 {\small ± 8.3} \\
ONLY-PRE+IFT 20K & 64.0 {\small ± 13.9} & 42.6 {\small ± 28.7} & 52.8 {\small ± 15.6} & 62.0 {\small ± 18.0} & 68.7 {\small ± 25.0} & 58.0 {\small ± 9.3} \\
ONLY-PRE+IFT 30K & 52.9 {\small ± 15.5} & 42.0 {\small ± 28.3} & 51.5 {\small ± 16.0} & 62.0 {\small ± 18.7} & 67.3 {\small ± 24.9} & 55.1 {\small ± 8.8} \\
ONLY-PRE+IFT 40K & 60.4 {\small ± 12.2} & 43.1 {\small ± 29.1} & 52.4 {\small ± 16.9} & 60.6 {\small ± 17.5} & 68.9 {\small ± 23.4} & 57.1 {\small ± 8.7} \\
ONLY-PRE+IFT 50K & 57.4 {\small ± 8.5} & 42.6 {\small ± 28.8} & 51.6 {\small ± 15.3} & 61.2 {\small ± 18.1} & 70.0 {\small ± 23.8} & 56.5 {\small ± 9.2} \\
ONLY-PRE+IFT 60K & 56.7 {\small ± 7.6} & 42.5 {\small ± 28.4} & 52.0 {\small ± 16.3} & 61.2 {\small ± 17.9} & 68.8 {\small ± 23.8} & 56.2 {\small ± 8.8} \\
ONLY-PRE+IFT 70K & 57.2 {\small ± 8.5} & 42.1 {\small ± 28.4} & 51.5 {\small ± 17.0} & 60.8 {\small ± 18.1} & 70.2 {\small ± 24.8} & 56.3 {\small ± 9.4} \\
ONLY-PRE+IFT 80K & 60.3 {\small ± 11.1} & 42.4 {\small ± 28.4} & 54.6 {\small ± 16.4} & 65.1 {\small ± 20.9} & 69.2 {\small ± 24.8} & 58.3 {\small ± 9.3} \\
ONLY-PRE+IFT 90K & 60.3 {\small ± 12.0} & 44.4 {\small ± 29.8} & 52.3 {\small ± 17.1} & 59.8 {\small ± 17.8} & 67.8 {\small ± 24.4} & 56.9 {\small ± 7.9} \\
\bottomrule
\end{tabular}
}
\label{tab:domain_adaptation_xl}
\end{table*}

\begin{table*}[h]
\centering
\caption{Flan-T5 XXL models with different domain adaptation strategies (amount of IFT data during continued pretraining). 1-IFT-to-X-PRE means that for every X pretraining examples we mix in one instruction example. ONLY-PRE means we did not mix in any instruction examples.}
\resizebox{\textwidth}{!}{
\begin{tabular}{lrrrrrr}
\toprule
\bf LLM & \bf Issue & \bf Rule & \bf Conclusion & \bf Interpretation & \bf Rhetorical & \bf LegalBench \\
\midrule
Baseline & 36.1 {\small ± 21.5} & 18.8 {\small ± 24.6} & 25.2 {\small ± 26.0} & 35.1 {\small ± 22.2} & 41.1 {\small ± 18.4} & 31.3 {\small ± 8.1} \\
IFT & 55.2 {\small ± 23.7} & 46.3 {\small ± 31.6} & 56.2 {\small ± 18.3} & 66.3 {\small ± 19.7} & 73.8 {\small ± 24.4} & 59.6 {\small ± 9.5} \\
\midrule
1-IFT-to-200-PRE+IFT 10K & 53.4 {\small ± 16.2} & 47.9 {\small ± 32.1} & 58.1 {\small ± 19.5} & 63.8 {\small ± 17.6} & 74.2 {\small ± 27.1} & 59.5 {\small ± 9.0} \\
1-IFT-to-200-PRE+IFT 20K & 53.6 {\small ± 3.7} & 48.9 {\small ± 32.9} & 58.8 {\small ± 18.7} & 65.3 {\small ± 17.5} & 72.0 {\small ± 25.5} & 59.7 {\small ± 8.2} \\
1-IFT-to-200-PRE+IFT 30K & 56.5 {\small ± 18.3} & 48.9 {\small ± 31.5} & 60.5 {\small ± 19.9} & 65.2 {\small ± 18.3} & 69.5 {\small ± 24.2} & 60.1 {\small ± 7.1} \\
1-IFT-to-200-PRE+IFT 40K & 58.3 {\small ± 20.2} & 47.3 {\small ± 30.8} & 57.9 {\small ± 19.1} & 65.6 {\small ± 18.2} & 71.3 {\small ± 24.1} & 60.1 {\small ± 8.1} \\
1-IFT-to-200-PRE+IFT 50K & 60.3 {\small ± 12.6} & 48.4 {\small ± 31.4} & 63.2 {\small ± 20.2} & 67.9 {\small ± 18.9} & 71.4 {\small ± 26.1} & 62.2 {\small ± 7.9} \\
1-IFT-to-200-PRE+IFT 60K & 58.6 {\small ± 20.5} & 48.5 {\small ± 31.5} & 60.9 {\small ± 20.7} & 67.5 {\small ± 19.9} & 71.0 {\small ± 24.7} & 61.3 {\small ± 7.8} \\
1-IFT-to-200-PRE+IFT 70K & 58.6 {\small ± 10.5} & 48.5 {\small ± 31.4} & 60.6 {\small ± 20.4} & 65.3 {\small ± 18.4} & 69.3 {\small ± 23.4} & 60.5 {\small ± 7.0} \\
1-IFT-to-200-PRE+IFT 80K & 53.7 {\small ± 16.4} & 47.8 {\small ± 30.8} & 58.8 {\small ± 18.2} & 63.7 {\small ± 17.7} & 71.3 {\small ± 25.7} & 59.1 {\small ± 8.1} \\
1-IFT-to-200-PRE+IFT 90K & 52.0 {\small ± 14.5} & 48.8 {\small ± 31.7} & 59.4 {\small ± 19.6} & 64.4 {\small ± 17.9} & 72.3 {\small ± 25.1} & 59.4 {\small ± 8.5} \\
\midrule
1-IFT-to-1000-PRE+IFT 10K & 41.1 {\small ± 24.2} & 45.9 {\small ± 30.3} & 58.2 {\small ± 18.4} & 65.5 {\small ± 20.2} & 68.8 {\small ± 25.2} & 55.9 {\small ± 10.8} \\
1-IFT-to-1000-PRE+IFT 20K & 47.7 {\small ± 24.8} & 48.0 {\small ± 31.1} & 60.3 {\small ± 20.3} & 67.2 {\small ± 19.7} & 70.3 {\small ± 23.8} & 58.7 {\small ± 9.4} \\
1-IFT-to-1000-PRE+IFT 30K & 40.3 {\small ± 28.4} & 45.5 {\small ± 29.6} & 62.3 {\small ± 21.1} & 67.8 {\small ± 21.1} & 69.3 {\small ± 22.6} & 57.0 {\small ± 11.9} \\
1-IFT-to-1000-PRE+IFT 40K & 44.2 {\small ± 27.4} & 46.7 {\small ± 29.9} & 61.9 {\small ± 21.9} & 68.6 {\small ± 20.7} & 71.2 {\small ± 24.9} & 58.5 {\small ± 11.1} \\
1-IFT-to-1000-PRE+IFT 50K & 49.7 {\small ± 25.2} & 49.1 {\small ± 33.1} & 55.5 {\small ± 19.2} & 68.2 {\small ± 19.8} & 71.4 {\small ± 24.3} & 58.8 {\small ± 9.3} \\
1-IFT-to-1000-PRE+IFT 60K & 44.9 {\small ± 22.0} & 47.6 {\small ± 30.7} & 57.9 {\small ± 19.4} & 69.7 {\small ± 21.1} & 72.1 {\small ± 26.0} & 58.5 {\small ± 11.1} \\
1-IFT-to-1000-PRE+IFT 70K & 40.6 {\small ± 25.0} & 48.1 {\small ± 31.2} & 60.5 {\small ± 20.0} & 68.2 {\small ± 20.5} & 72.5 {\small ± 24.4} & 58.0 {\small ± 12.0} \\
1-IFT-to-1000-PRE+IFT 80K & 53.8 {\small ± 23.7} & 47.9 {\small ± 32.4} & 53.5 {\small ± 17.5} & 67.1 {\small ± 19.3} & 71.8 {\small ± 25.9} & 58.8 {\small ± 9.1} \\
1-IFT-to-1000-PRE+IFT 90K & 47.6 {\small ± 23.5} & 47.1 {\small ± 30.5} & 60.1 {\small ± 18.9} & 65.1 {\small ± 24.3} & 70.3 {\small ± 23.5} & 58.0 {\small ± 9.3} \\
\midrule
1-IFT-to-10000-PRE+IFT 10K & 49.8 {\small ± 13.6} & 46.6 {\small ± 30.0} & 59.0 {\small ± 16.6} & 64.6 {\small ± 19.3} & 72.6 {\small ± 24.7} & 58.5 {\small ± 9.5} \\
1-IFT-to-10000-PRE+IFT 20K & 45.2 {\small ± 27.4} & 46.3 {\small ± 31.2} & 58.8 {\small ± 20.1} & 68.1 {\small ± 19.0} & 71.7 {\small ± 24.1} & 58.0 {\small ± 10.9} \\
1-IFT-to-10000-PRE+IFT 30K & 46.8 {\small ± 24.6} & 46.0 {\small ± 29.6} & 62.6 {\small ± 18.4} & 66.1 {\small ± 18.1} & 72.1 {\small ± 25.3} & 58.7 {\small ± 10.5} \\
1-IFT-to-10000-PRE+IFT 40K & 56.8 {\small ± 24.5} & 46.9 {\small ± 30.4} & 59.1 {\small ± 19.3} & 68.3 {\small ± 21.1} & 72.2 {\small ± 26.2} & 60.7 {\small ± 8.9} \\
1-IFT-to-10000-PRE+IFT 50K & 54.5 {\small ± 28.7} & 43.1 {\small ± 28.1} & 62.2 {\small ± 19.8} & 64.2 {\small ± 19.1} & 70.2 {\small ± 24.3} & 58.8 {\small ± 9.3} \\
1-IFT-to-10000-PRE+IFT 60K & 52.0 {\small ± 16.0} & 42.0 {\small ± 28.7} & 60.3 {\small ± 17.4} & 65.7 {\small ± 19.6} & 71.3 {\small ± 24.7} & 58.2 {\small ± 10.3} \\
1-IFT-to-10000-PRE+IFT 70K & 52.2 {\small ± 14.7} & 47.4 {\small ± 30.8} & 59.2 {\small ± 18.3} & 66.6 {\small ± 18.5} & 70.0 {\small ± 24.1} & 59.1 {\small ± 8.5} \\
1-IFT-to-10000-PRE+IFT 80K & 56.5 {\small ± 18.5} & 44.9 {\small ± 28.9} & 59.7 {\small ± 17.2} & 65.3 {\small ± 17.7} & 72.3 {\small ± 25.6} & 59.7 {\small ± 9.1} \\
1-IFT-to-10000-PRE+IFT 90K & 45.0 {\small ± 17.4} & 41.5 {\small ± 27.3} & 56.3 {\small ± 16.3} & 66.3 {\small ± 18.5} & 72.1 {\small ± 25.7} & 56.2 {\small ± 11.8} \\
\midrule
ONLY-PRE+IFT 10K & 49.2 {\small ± 24.4} & 47.1 {\small ± 30.4} & 62.0 {\small ± 20.3} & 66.9 {\small ± 20.4} & 71.7 {\small ± 25.1} & 59.4 {\small ± 9.7} \\
ONLY-PRE+IFT 20K & 35.6 {\small ± 24.0} & 46.2 {\small ± 30.0} & 56.3 {\small ± 17.9} & 62.3 {\small ± 18.4} & 68.6 {\small ± 24.2} & 53.8 {\small ± 11.7} \\
ONLY-PRE+IFT 30K & 46.3 {\small ± 28.4} & 45.7 {\small ± 29.3} & 56.1 {\small ± 18.5} & 67.7 {\small ± 19.9} & 72.1 {\small ± 25.6} & 57.6 {\small ± 10.8} \\
ONLY-PRE+IFT 40K & 48.8 {\small ± 30.3} & 45.7 {\small ± 29.5} & 56.6 {\small ± 18.0} & 68.1 {\small ± 20.0} & 71.6 {\small ± 26.3} & 58.1 {\small ± 10.2} \\
ONLY-PRE+IFT 50K & 47.5 {\small ± 24.9} & 47.1 {\small ± 30.2} & 53.5 {\small ± 16.2} & 67.1 {\small ± 19.5} & 71.8 {\small ± 25.4} & 57.4 {\small ± 10.2} \\
ONLY-PRE+IFT 60K & 33.2 {\small ± 23.3} & 47.8 {\small ± 30.7} & 55.0 {\small ± 17.9} & 63.1 {\small ± 19.7} & 69.3 {\small ± 25.0} & 53.7 {\small ± 12.6} \\
ONLY-PRE+IFT 70K & 42.7 {\small ± 25.9} & 47.2 {\small ± 30.5} & 55.9 {\small ± 19.4} & 60.7 {\small ± 17.5} & 68.0 {\small ± 23.8} & 54.9 {\small ± 9.1} \\
ONLY-PRE+IFT 80K & 43.7 {\small ± 25.8} & 46.3 {\small ± 29.9} & 55.8 {\small ± 17.1} & 64.8 {\small ± 18.7} & 71.8 {\small ± 25.9} & 56.5 {\small ± 10.7} \\
ONLY-PRE+IFT 90K & 55.3 {\small ± 16.9} & 45.2 {\small ± 28.9} & 60.0 {\small ± 17.0} & 64.9 {\small ± 20.0} & 69.0 {\small ± 24.3} & 58.9 {\small ± 8.2} \\
\bottomrule
\end{tabular}
}
\label{tab:domain_adaptation_xxl}
\end{table*}

\begin{figure*}[ht]
\centering
\begin{subfigure}[b]{0.49\textwidth}
    \centering
    \includegraphics[width=\columnwidth]{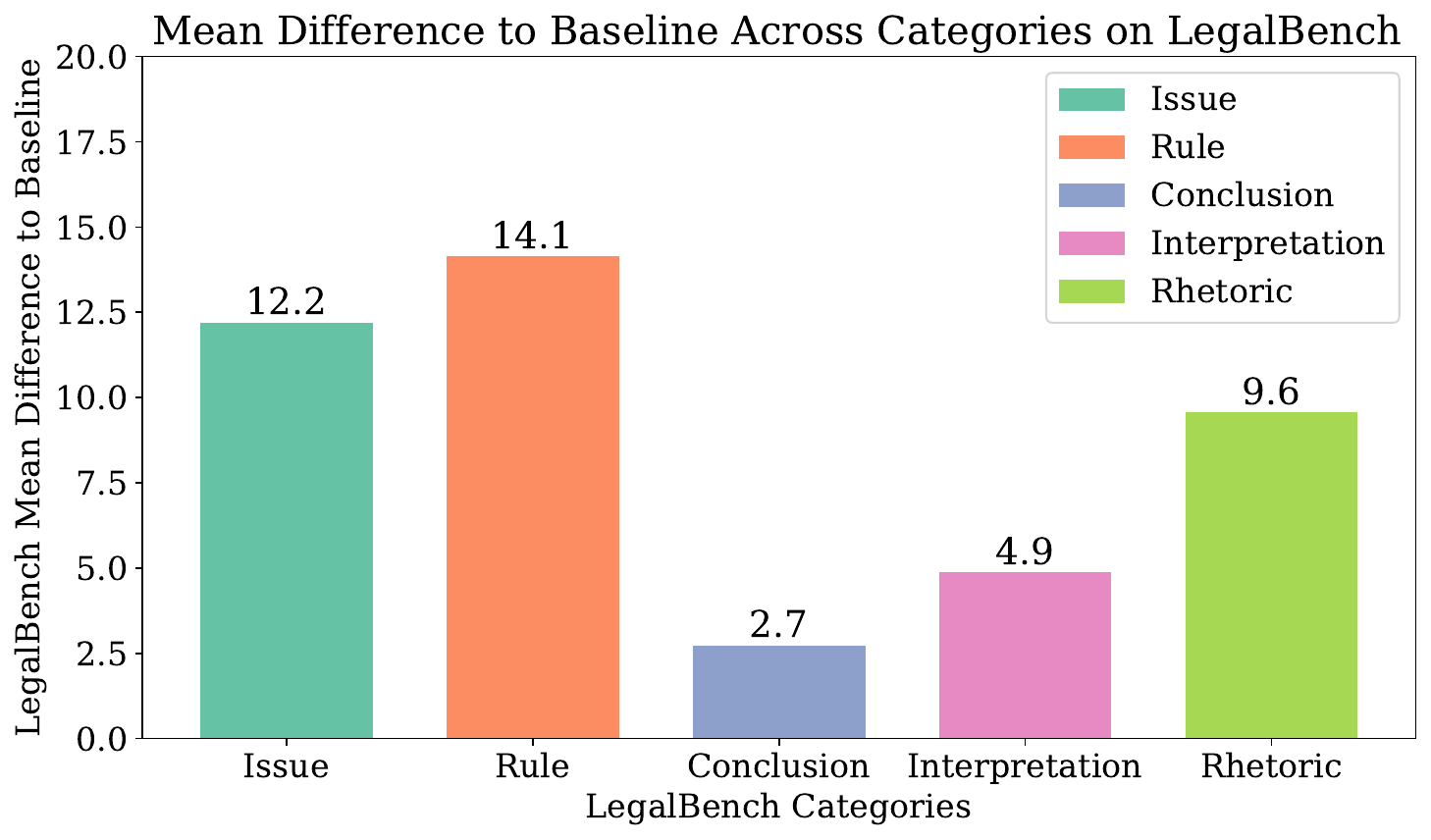}
    \caption{Dataset Overlap}
    \label{fig:difference_to_baseline_categories_lb_held_out_dataset}
\end{subfigure}%
\hfill 
\begin{subfigure}[b]{0.49\textwidth}
    \centering
    \includegraphics[width=\columnwidth]{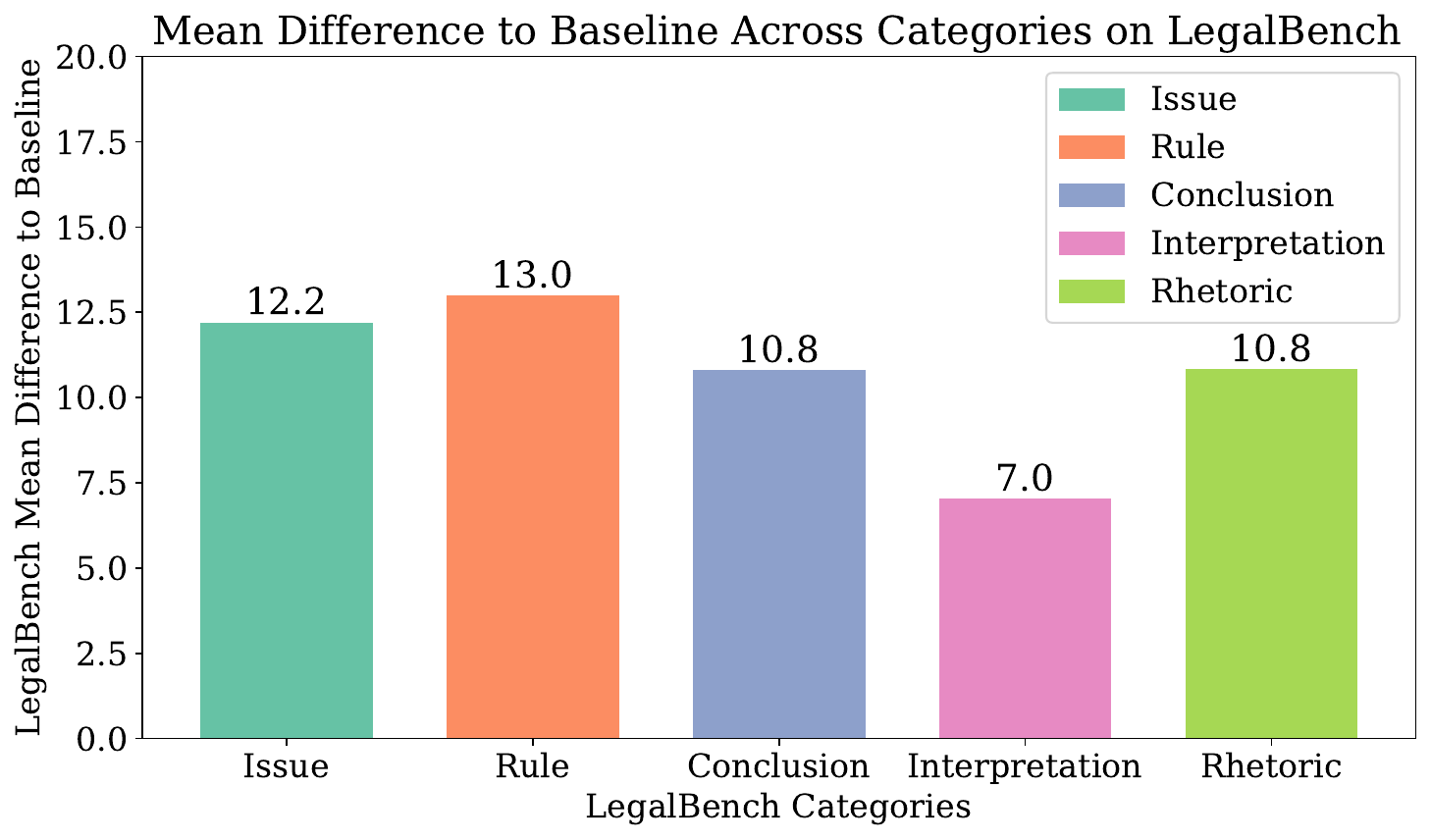}
    \caption{Task Overlap}
    \label{fig:difference_to_baseline_categories_lb_held_out_task}
\end{subfigure}
\caption{Difference to the baseline for the XL model across categories on LegalBench with dataset and task overlap held out respectively.}
\label{fig:difference_to_baseline_categories_lb_heldout}
\end{figure*}

\end{document}